\documentclass[]{imag-ms-template}

\usepackage{caption}
\usepackage{subcaption}
\usepackage{mleftright}
\usepackage{float}
\usepackage{tikz}
\usepackage{xcolor}
\usepackage{colortbl}
\usepackage{microtype}

\usepackage{lineno}

\definecolor{linkblue}{RGB}{0,90,160}
\hypersetup{colorlinks=true, urlcolor=linkblue, citecolor=black,
  linkcolor=black}

\newcommand{\weblink}[2]{%
  \href{#1}{\begingroup\urlstyle{same}\nolinkurl{#2}\endgroup}}

\makeatletter
\g@addto@macro\normalsize{%
  \setlength{\abovedisplayskip}{6pt plus 3pt minus 2pt}%
  \setlength{\belowdisplayskip}{6pt plus 3pt minus 2pt}%
  \setlength{\abovedisplayshortskip}{3pt plus 3pt}%
  \setlength{\belowdisplayshortskip}{4pt plus 3pt minus 2pt}}
\makeatother

\usepackage{titlesec}
\titlespacing*{\section}{0pt}{1.6ex plus .6ex minus .2ex}{0.6ex plus .2ex}
\titlespacing*{\subsection}{0pt}{1.3ex plus .5ex minus .2ex}{0.3ex plus .2ex}
\titlespacing*{\subsubsection}{0pt}{1.1ex plus .4ex minus .2ex}{0.2ex plus .2ex}

\makeatletter
\patchcmd{\@maketitle}{\vskip 2em}{\vskip 0.2em}{}{}
\patchcmd{\@maketitle}{\vskip 1.5em}{\vskip 0.8em}{}{}
\patchcmd{\@maketitle}{\vskip 1em}{\vskip 0.4em}{}{}
\patchcmd{\@maketitle}{\vskip 1.5em}{\vskip 0.8em}{}{}
\makeatother

\newcommand\numberthis{\addtocounter{equation}{1}\tag{\theequation}}
\DeclareMathOperator{\Tr}{Tr}
\DeclareMathOperator{\LLR}{LLR}

\title{GLOW: General Linear Models Optimized with Ward's Method}
\author{Matt Higger$^{1\ast}$\\
{\small $^{1}$Khoury College of Computer Sciences, Northeastern University,}\\
{\small Boston, MA 02115, USA}\\
{\small $^\ast$Correspondence: mhigger@ccs.neu.edu}\\
{\small ORCID: \href{https://orcid.org/0000-0001-6384-9114}{0000-0001-6384-9114}}
}

\begin{document}

\maketitle

\keywords{general linear model, MANCOVA, permutation inference, family-wise error rate, Ward's method, hierarchical segmentation}

\begin{abstract}
	Identifying brain regions whose imaging features correlate with variables of interest (e.g.\ diagnosis, age) is a central goal of neuroimaging, usually pursued with the General Linear Model (GLM).
	Region-Based Analysis dilutes or splits any effect which does not fill exactly one atlas region.
	Voxel-Based Analysis (VBA), with Cluster Extent Thresholding (CET) and Threshold-Free Cluster Enhancement (TFCE) built on it, fits a GLM per voxel and groups adjacent significant voxels.
	But neighboring voxels can share a statistic without sharing a GLM, and a per-voxel statistic cannot tell them apart.
	GLOW (General Linear Models Optimized with Ward's method) instead builds a hierarchy of nested regions by greedily minimizing GLM error, which we show is a special case of Ward's clustering.
	Maximum-statistic (Westfall-Young) permutation controls family-wise error over all regions.
	Significance spreads through the hierarchy, to an effect's nested fragments and the regions containing it.
	GLOW's final step seeks to prune away these distractions to output whole, pure effect extent estimates.
	We validate GLOW by planting synthetic effects of known extent and varying strength in diffusion images from the Human Connectome Project (HCP) and in white noise.
	Pruning delivers wholeness on spatially consistent effects: GLOW returns one as a single region on $57\%$ of HCP trials against the best baseline's $35\%$; the baselines recover comparable volume, but in many pure pieces.
	The cost is positive predictive value (PPV): GLOW is more sensitive (by up to $0.29$), but its regions overrun the effect, mostly because no region of the hierarchy isolates the effect rather than because of the pruning.
	On HCP, GLOW's average Dice gains $0.059$ over the best baseline at weaker effects and gives back $0.005$ at stronger ones.
	On white noise, GLOW trails severely, as much as $0.61$: its hierarchy holds the effect, but its permutation test does not certify it.
	Its runtime grows roughly linearly with voxel count, reaching about four and a half times that of the voxel-wise methods on the full HCP volume.
	The PPV loss may trace to the assumption of spatially independent residuals.
\end{abstract}

\section{Introduction}\label{sec:intro}
A fundamental aim of neuroimaging research is to identify how independent variables (e.g.\ age, diagnosis, or behavioral measures) influence brain structure \parencite{friston1995statistical,ashburner2000voxel}.
For example, white-matter microstructure, as measured by fractional anisotropy (FA) or mean diffusivity (MD), may differ between clinical groups or correlate with behavioral scores \parencite{smith2006tbss}.
Such a study images a set of individuals whose independent variable varies, then registers those images into a common anatomical space \parencite{jenkinson2002improved,avants2011reproducible}.
Throughout this work, we assume that registration has already been done.

The General Linear Model (GLM) provides a unified framework for testing whether image intensities correlate with independent variables after controlling for confounds \parencite{friston1995statistical,worsley2002general}.
By encoding hypotheses in a design-matrix/contrast framework, the GLM supports $t$-tests, ANOVA, multiple regression, and factorial designs, with nuisance regressors explicitly adjusting for covariates such as age, sex, or acquisition site.

Before applying a GLM, one must decide how to group voxels for analysis.
Region-Based Analyses (RBA) define regions a priori, typically from anatomical atlases \parencite{tzourio2002automated,desikan2006automated,Glasser2016MMP}.
Each region's GLM tests whether an effect exists uniformly across its voxels.
Statistical power suffers in two complementary ways: including non-effect voxels in a region dilutes the signal, while splitting a single effect across atlas boundaries fragments it \parencite{eickhoff2018parcellations}.
Both errors are to be expected: atlas boundaries track anatomy, not the unknown extent of some effect.

Voxel-Based Analyses (VBA) sidestep atlas dependence by fitting a GLM at every voxel and gluing adjacent, significant voxels together post hoc \parencite{ashburner2000voxel,friston1995statistical}.
A central challenge for VBA is incorporating spatial scale: a lone island of significance is more likely to be a false positive than a large, contiguous continent of significant voxels.
Cluster Extent Thresholding (CET) \parencite{friston1994assessing,forman1995improved} and Threshold-Free Cluster Enhancement (TFCE) \parencite{smith2009threshold} act as spatial regularization steps on top of VBA, favoring continents over islands.
CET thresholds the statistic map and keeps those clusters of suprathreshold voxels whose extent is unlikely under the null.
TFCE sidesteps the choice of any one threshold by using all of them: each voxel's statistic is enhanced in proportion to the extent of the suprathreshold clusters containing it, accumulated over every threshold.
Because each voxel is tested on its own, the resulting map must be corrected for a massive number of comparisons.
The family-wise error rate (FWER) \parencite{nichols2003controlling} is controlled either parametrically, via Random Field Theory \parencite{worsley1996unified}, or nonparametrically, via permutation \parencite{nichols2002nonparametric,winkler2014permutation}.
Multivariate data fit the same voxel-wise template: PALM tests several imaging features jointly at each voxel, via either classical multivariate statistics or non-parametric combination \parencite{winkler2016npc}, still leaving the single voxel as the unit of inference.

VBA's first aggregation across voxels comes before any test: each image is smoothed with a Gaussian kernel, so every voxel's statistic already averages its neighbors.
By the matched filter theorem \parencite{turin1960introduction}, smoothing best detects a signal when the kernel matches the signal's own extent: wide effects want a wide kernel, narrow effects a narrow one \parencite{worsley1996searching,friston2004experimental}.
That extent is not known a priori, so the analyst fixes one width in advance for the whole image, and conclusions can turn on the choice: kernels of 0 to 16~mm led one diffusion data set to four different conclusions \parencite{jones2005effect}.
We smooth each voxel-wise baseline at the width which serves it best on our data (Appendix~\ref{sec:stat_selection}).

\begin{figure}[htbp]
	\centering
	\includegraphics[width=\textwidth]{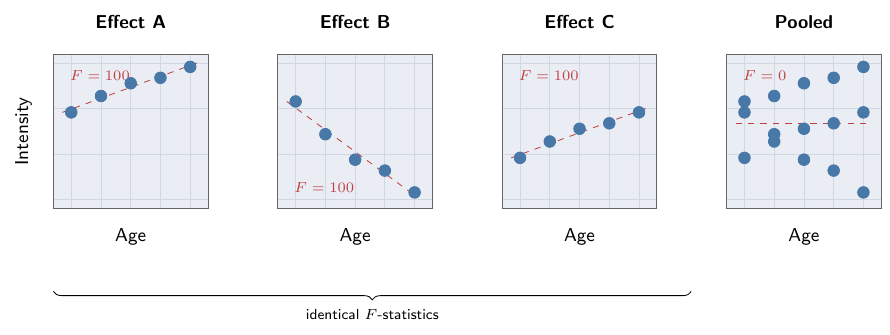}
	\caption{Toy example: three voxels whose effects (A, B, C) are distinct yet share the same $F$-statistic. Pooling across voxels reveals their heterogeneity.}
	\label{fig:same_f}
\end{figure}

VBA and the cluster methods built on it share a clear advantage over RBA: their output regions can mold to the contours of effects present in the data.
But pooling per-voxel statistics, as CET and TFCE do, or blurring the images each voxel's model is fit to, as smoothing does, is not the same as fitting one model across voxels.
A per-voxel statistic is all VBA ever sees of a voxel's GLM, and there is plenty of wiggle room among GLMs which share an $F$-statistic (Fig~\ref{fig:same_f}).

Consider the three neighboring voxels of Fig~\ref{fig:same_f}, whose effects A, B, and C are clearly distinct yet produce an identical per-voxel statistic.
Seeing three identical statistics, VBA is encouraged to glue the voxels together into a single output.
The statistic is a bottleneck: whatever passes through it cannot tell VBA whether neighboring voxels agree on a model, so VBA can neither reward that agreement nor penalize its absence.

The bottleneck costs detection power, not only interpretation.
Under the null, where imaging features carry no correlation with the independent variables, $n$ contiguous voxels which each reach $F = 100$ under one shared effect $A$ are rarer than $n$ contiguous voxels which each reach $F = 100$ under whatever model they please; the first event is contained in the second.
A per-voxel test draws its null from the second, looser event, so the permuted images are free to cobble together unrelated GLMs and compete with the observed statistic.
Asking the voxels to agree on a model narrows what the null can offer, which is the power GLOW is after.

Cluster-wise inference has known weaknesses: \textcite{woo2014cluster} and \textcite{eklund2016cluster} show that it can inflate false-positive rates and pool heterogeneous signals.
A more principled remedy comes from All-Resolutions Inference (ARI) and its true-discovery-proportion descendants, which report a lower bound on the fraction of truly active voxels holding simultaneously for any region the analyst selects \parencite{Rosenblatt2018ARI,Blain2022Notip}.
Yet the bound is a proportion, not a partition: for whatever region the analyst queries it reports how many voxels are active, never which ones.

Data-driven parcellation approaches, including spatially constrained clustering \parencite{craddock2012whole,thirion2014which,schaefer2018local,michel2012}, address the atlas-dependence problem but do not optimize regions specifically for GLM fit \parencite{eickhoff2018parcellations}.\footnote{Michel et al.\ are a partial exception: their tree is built by unsupervised connectivity-constrained Ward clustering, but the variable of interest enters only when \emph{cutting} the tree (selecting the parcellation whose averages best decode the target).  We return to this in Section~\ref{ssec:prune}.}
Our objective is to construct regions whose voxels follow one spatially homogeneous GLM.
GLOW (General Linear Models Optimized with Ward's method) does so by building a hierarchical segmentation which greedily minimizes each region's GLM mean-squared error, via Ward's agglomerative clustering \parencite{ward1963hierarchical}; in practice it measures that error only along the variables of interest (Section~\ref{ssec:method:hier_seg}).
The significance of all regions is evaluated with the same permutation machinery used for voxel-wise inference: Freedman--Lane residualization \parencite{freedman1983nonstochastic} and Westfall-Young max-statistic FWER \parencite{westfall1993resampling}, as set out by \textcite{winkler2014permutation}.
The only change is that this machinery is applied to GLOW's data-derived regions rather than to fixed voxels or threshold-formed clusters.
GLOW is multivariate as well: it analyzes several imaging features jointly (e.g.\ fractional anisotropy and mean diffusivity) within the GLM/MANCOVA framework \parencite{rencher2012methods}, as PALM's multivariate tests do at the voxel level \parencite{winkler2016npc}.

We validate GLOW by imposing synthetic effects onto image sets of known ground truth and comparing detection performance against VBA, CET and TFCE.
One image set honors our statistical assumptions (white Gaussian noise); the other carries the spatial correlations of real data (diffusion images from the Human Connectome Project \parencite{vanessen2013wu}).
In the remainder of the paper we analyze the GLM in the special case here where every voxel shares one design matrix (Section~\ref{sec:analysis}), describe GLOW's method (Section~\ref{sec:method}), present experiments and results (Sections~\ref{sec:experiment} and~\ref{sec:results}), and discuss how spatial dependence and region selection limit GLOW (Section~\ref{sec:discussion}).

\section{Problem \& Analysis} \label{sec:analysis}

\subsection{Problem Statement: Multiple-Hypothesis GLM} \label{ssec:problem}

Let $y_{v, s} \in \mathbb{R}^b$ hold the imaging features of voxel $v$ in image $s$, one of $N$ images registered into a common anatomical space.
Each (whole) image carries a vector of independent variables $x_s \in \mathbb{R}^a$ (e.g.\ diagnostic labels, demographic covariates, or clinical scores).

Among these $a$ independent variables, $a' < a$ are covariates (nuisance variables) whose correlation with the imaging features is not of interest; the remaining $a - a'$ are the variables of interest.
We seek spatially contiguous effect regions $r$, each a set of voxels, in which $x_s$ predicts $y_{v, s}$ through one GLM shared by every $v \in r$:

\begin{equation}
	y_{v, s} = x_s \beta_r + \epsilon,
	\label{eqn:regression_simple}
\end{equation}

where $\beta_r \in \mathbb{R}^{a \times b}$ is the matrix of regression coefficients for region $r$, and $\epsilon$ is additive white Gaussian noise.
We test the alternative $H_1$, that some entry of $\beta_r$ is non-zero in the rows belonging to the $a - a'$ variables of interest, against the null $H_0$ that every such entry is zero.

Finding the voxels which reject $H_0$ is only half the job; we also want them collected into output regions holding all, and only, those voxels governed by a single homogeneous GLM.
In the language of Metrics Reloaded~\parencite{maierhein2024metrics}, this is an instance segmentation problem: a GLM's extent should be neither fragmented across several output regions nor conflated with neighboring voxels.
We quantify the two errors with Homogeneity and Completeness scores~\parencite{rosenberg2007vmeasure}:
\begin{equation}
	\text{completeness} = 1 - \frac{H(R \mid T)}{H(R)} = \frac{I(R, T)}{H(R)},
	\qquad
	\text{homogeneity} = 1 - \frac{H(T \mid R)}{H(T)} = \frac{I(R, T)}{H(T)}.
	\label{eqn:hom_com}
\end{equation}
Here $H$ and $I$ are entropy and mutual information, $R$ is the index of an output region (the output regions being pairwise disjoint) and $T$ is the ground truth effect label ($T = 0$ outside every planted effect, $T = k$ inside effect $k$).
In computing them we assume a uniform distribution of voxels over the union of the output regions.\footnote{Weighting voxels rather than matched regions frees these scores from the one-to-one correspondence between output and ground truth regions that Panoptic Quality requires; the discussion of \textcite{maierhein2024metrics} places problems where that correspondence is infeasible outside their recommendations, pointing instead to the Variation of Information \parencite{meila2007comparing}, a function of the completeness and homogeneity metrics above.}
Completeness is the percentage of the uncertainty in $R$ which observing $T$ resolves; it reaches $100\%$ when each whole effect sits inside a single output region.
Homogeneity is the percentage of the uncertainty in $T$ which observing $R$ resolves; it reaches $100\%$ when no output region mixes two effects.
Both are computed over detected voxels alone, so neither penalizes effect volume that went undetected.
We therefore report them alongside sensitivity and PPV.

\subsection{GLM/MANCOVA via QR Factorization (single-voxel)}
We first review\footnote{see the ``QR Factorization'' section of ``Least Squares Problem'' in \textcite{trefethen1997numerical}} GLM regression through a QR factorization of the design matrix $X \in \mathbb{R}^{N \times a}$; this single-voxel case lays the foundation for the multi-voxel extension which follows.

We perform a QR factorization of the design matrix $X$:
\begin{equation}
	X = QR
	\label{eqn:x_equal_qr}
\end{equation}
where $Q \in \mathbb{R}^{N \times a}$ has orthonormal columns (i.e., $Q^T Q = I_a$), and $R \in \mathbb{R}^{a \times a}$ is upper triangular.
Because $R$ is upper triangular and invertible (as $X$ has full column rank), the first $k$ columns of $X$ and the first $k$ columns of $Q$ span the same subspace.
Before computing this QR factorization, we re-index the columns of $X$ so that covariates appear first, followed by the independent variables of interest.
Taken together, these facts let us partition $Q$:
\begin{equation}
	Q = \begin{bmatrix} Q_0 \mid Q_1 \end{bmatrix}
	\label{eqn:q_cols_012}
\end{equation}
where:
\begin{itemize}
	\item $Q_0 \in \mathbb{R}^{N \times a'}$ spans the space of covariates. Components of $Y$ in this span can be entirely attributed to nuisance variables.
	\item $Q_1 \in \mathbb{R}^{N \times (a - a')}$ spans the independent variables of interest, orthogonalized with respect to covariates. Components of $Y$ in this span are explained by independent variables of interest while also being orthogonal to all nuisance variables.
	\item Every direction orthogonal to both spans is residual: no linear combination of the independent variables reaches it.
\end{itemize}

In the univariate case ($b = 1$), hypothesis testing is typically performed with $F$-statistics \parencite{kutner2005applied}.
With multiple imaging features ($b > 1$) the problem generalizes to MANCOVA, but the spirit of the analysis is unchanged: a GLM's fit is the ratio of the variance in $y$ it explains (the hypothesis matrix $H$) to the variance left over (the error, or residual, matrix $E$).
We define $E$ precisely:
\begin{equation}
	E \equiv (Y_v - X \beta)^T(Y_v - X \beta) = Y_v^T (I_N - Q Q^T) Y_v,
	\label{eqn:e_single_voxel}
\end{equation}
where $Y_v \in \mathbb{R}^{N \times b}$ are the imaging features of voxel $v$.
The second equality is seen by observing that when $\beta$ is chosen to minimize the mean-squared-error, $X \beta$ is a projection of $Y_v$ into the column space of $X$: $X \beta = Q Q^T Y_v$.
Equivalently, the residual is a projection into the orthogonal complement of $X$: $Y_v - X \beta = (I_N - Q Q^T) Y_v$.

The hypothesis matrix $H$ captures the portion of the covariance in $Y_v$ explained by independent variables of interest, while being orthogonal to all nuisance variables:
\begin{equation}
	H = Y_v^T Q_1 Q_1^T Y_v
	\label{eqn:h_single_vox}
\end{equation}

A voxel with a strong effect yields a small $E$ and a large $H$: much of the variance in $Y$ is explained by the independent variables of interest (the last $a - a'$ columns of $X$).
MANCOVA offers four generalizations of the $F$-statistic (Wilks' Lambda, Pillai's Trace, Hotelling--Lawley Trace and Roy's Largest Root), each a function of $E$ and $H$ \parencite{rencher2012methods}.

GLOW instead scores a region by an estimate of its log-likelihood ratio (LLR), drawing the null distribution from permutation (Section~\ref{ssec:method:sig_test}) rather than from parametric theory.
To derive it, take the residuals to be independent and identically normal.
Appendix~\ref{sec:normal_identity} then shows that the log-likelihood of $N|r|$ residuals is $- \frac{N|r|}{2} \log |\hat{\Sigma}|$ plus constants depending only on $N$, $b$ and $|r|$, where $\hat{\Sigma}$ is their maximum-likelihood covariance estimate and $|r|$ is the number of voxels in the region.
The residuals under $H_1$, the full model, give $\hat{\Sigma} = E / (N|r|)$, while under $H_0$, the reduced model, they give $\hat{\Sigma} = (E + H) / (N|r|)$.
Subtracting the two log-likelihoods cancels the shared constants, along with the $(N|r|)^{-1}$ inside each determinant:
\begin{equation}
	\text{LLR} = \frac{N|r|}{2} \log \frac{|E + H|}{|E|} \label{eqn:llr}
\end{equation}
Here $|\cdot|$ denotes a matrix determinant; the region size $|r|$ uses the same bars for cardinality.
Equation~(\ref{eqn:llr}) is an estimate because the true covariance $\Sigma$ is unknown: each model's log-likelihood is evaluated at its maximum-likelihood estimate $\hat{\Sigma}$ instead of at $\Sigma$, so Equation~(\ref{eqn:llr}) approaches the true log-likelihood ratio only as $\hat{\Sigma} \to \Sigma$.

The independence assumption, of course, is rarely valid in practice: residuals depend on one another wherever an image's spatial autocorrelation is non-zero.
This modeling error haunts our results.

\subsection{GLM/MANCOVA via QR Factorization (multi-voxel)}
Extending $E$ and $H$ to the multi-voxel case yields insight and, as seen later, desperately needed computational savings.
The theme throughout is exploiting redundancy in the block structure of the multi-voxel analogues of $X$ and $Y$:
\begin{equation}
	X' = \begin{bmatrix}
		X      \\
		X      \\
		\vdots \\
		X
	\end{bmatrix}
	\in \mathbb{R}^{N |r| \times a}, \quad Y' = \begin{bmatrix}
		Y_1 \\ Y_2 \\ \vdots \\ Y_{|r|}
	\end{bmatrix} \in \mathbb{R}^{N |r| \times b}
	\label{eqn:xy_stacked}
\end{equation}

where region $r$ is a set of voxels $v$, each with imaging features $Y_v \in \mathbb{R}^{N \times b}$.
As a first example, the minimum-MSE mapping $\beta$ for a region follows from the spatial average $\bar{Y}_r = \frac{1}{|r|} \sum_{v \in r} Y_v$ alone, never touching the full $Y'$:

\begin{align*}
	\beta
	 & = (X'^T X')^{-1} X'^T Y'
	\\
	 & = ([X^T \hdots ]
	\begin{bmatrix}
		X \\ \vdots
	\end{bmatrix}
	)^{-1} [X^T \hdots ]
	\begin{bmatrix}
		Y_1 \\ \vdots
	\end{bmatrix}
	\\
	 & = (|r| X^T X)^{-1} X^T  (\sum_{v \in r} Y_v)
	\\
	 & = (X^T X)^{-1}  X^T  \bar{Y}_r  \numberthis
	\label{eqn:beta_is_volume_weighted}
\end{align*}

Extending $E$ and $H$ to a region needs a QR factorization of $X'$.
Thanks again to the redundant structure of $X'$, we can nearly re-use the one we have for $X$:
\begin{equation}
	Q' = \frac{1}{\sqrt{|r|}} \begin{bmatrix}
		Q \\
		Q \\
		\vdots
	\end{bmatrix} \in \mathbb{R}^{N |r| \times a}, \quad R' = \sqrt{|r|} R
	\label{eqn:q_r_multi_voxel}
\end{equation}
which has $X' = Q' R'$, upper triangular $R'$ and $Q'^T Q' = I_a$ as required.
Before computing $E$ it helps to define the spatial (image-pooled) covariance matrix $S \in \mathbb{R}^{b \times b}$, which measures how $Y_v$ varies across the voxels of $r$:
\begin{align*}
	S & = \sum_{v \in r} (Y_v - \bar{Y}_r)^T (Y_v - \bar{Y}_r)
	\\
	  & = Y'^T Y' - |r| \bar{Y}_r^T \bar{Y}_r
	\\
	  & = Y'^T (I_{N|r|} - |r| C^T C) Y'  \numberthis
	\label{eqn:sigma}
\end{align*}
where $I_{N|r|}$ is the $N|r| \times N|r|$ identity matrix and $C = \frac{1}{|r|} [I_N, I_N, \hdots]$ is the centering matrix which gives $\bar{Y}_r = C Y'$.
We leave $S$ un-normalized, for consistency with $E$ and $H$.
The multi-voxel form of $E$ now follows from (\ref{eqn:e_single_voxel}):
\begin{align*}
	E & = Y'^T (I_{N|r|} - Q' Q'^T) Y'    \numberthis \label{eqn:e_multi_voxel_first_line}  \\
	  & = Y'^T (I_{N|r|} - |r| C^T C + |r| C^T C - Q' Q'^T) Y'
	\\
	  & =  S + \frac{1}{|r|} Y'^T  \begin{bmatrix}
		                               I_N - Q Q^T & \cdots & I_N - Q Q^T \\
		                               \vdots      & \ddots & \vdots      \\
		                               I_N - Q Q^T & \cdots & I_N - Q Q^T
	                               \end{bmatrix} Y'                       \\
	  & = S + |r| \bar{Y}_r^T (I_N - Q Q^T) \bar{Y}_r \numberthis \label{eqn:e_multi_voxel}
\end{align*}
where we use (\ref{eqn:sigma}) in the third equality.

Equation~(\ref{eqn:e_multi_voxel}) splits the error into two sources.
The first, $|r| \bar{Y}_r^T (I_N - Q Q^T) \bar{Y}_r$, is a regression error on the spatial average: it closely resembles the single-voxel $E$ of (\ref{eqn:e_single_voxel}) and measures how far $\bar{Y}_r$ strays in directions $X$ cannot explain.
The second, the spatial variance $S$, is surprising at first because it does not depend on $X$ at all.
Every voxel of region $r$ in image $s$ shares one estimate $x_s \beta$, so the observations $y_{v,s}$ can only sit close to that estimate by sitting close to each other: variation across space contributes exclusively to $E$.

The multi-voxel Hypothesis Matrix can be derived from (\ref{eqn:h_single_vox}):
\begin{equation}
	H = Y'^T Q_1' Q_1'^T Y' = |r| \bar{Y}_r^T Q_1 Q_1^T \bar{Y}_r
	\label{eqn:h_multi_vox}
\end{equation}

\section{Method} \label{sec:method}
\subsection{Overview}\label{ssec:overview}
\begin{figure}[htbp]
	\centering
	\includegraphics[width=\textwidth]{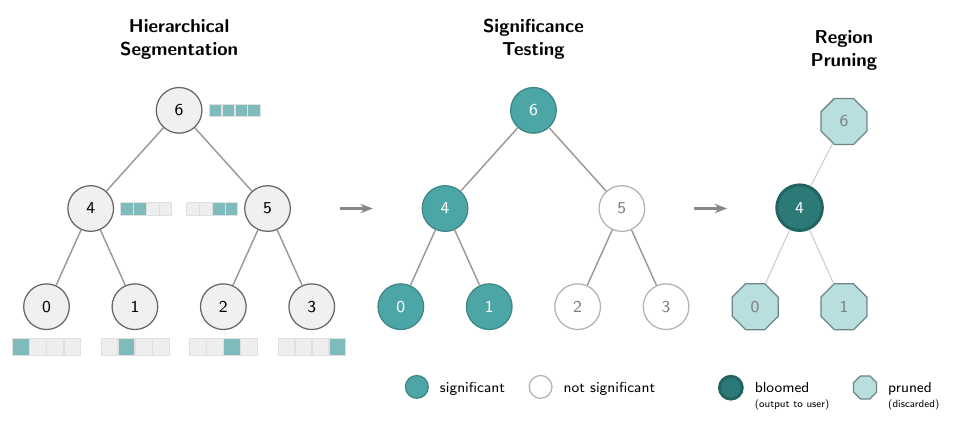}
	\caption{GLOW operating on a $1 \times 4$ array of voxels.  Regions 0 to 3 are single voxels, regions 4 to 6 their merges.  GLOW estimates that region 4 (voxels 0 and 1) contains an effect.  Regions 0, 1, 4 and 6 were all found significant, but only region 4 is output to the user; the rest are pruned away, their significance explained by their intersection with region 4.  The viewer at \weblink{https://glow-viewer.web.app/}{glow-viewer.web.app} is another helpful entry point.}
	\label{fig:summary}
\end{figure}
GLOW first segments the image into candidate regions, greedily minimizing a variant of the GLM fit's mean squared error (Section~\ref{ssec:method:hier_seg}).
It then tests each region with established permutation methods, discarding any whose fit the null hypothesis readily explains (Section~\ref{ssec:method:sig_test}).
A final pruning step (Section~\ref{ssec:prune}) sifts the surviving regions for those which contain all, and only, the voxels of one effect's extent.

\subsection{Hierarchical Segmentation} \label{ssec:method:hier_seg}
We seek regions which show a convincingly non-zero correlation between $Y$ and the independent variables of interest.
To that end GLOW uses agglomerative clustering, repeatedly merging the neighboring pair of regions whose GLMs are most alike.
Starting from one voxel per region, it merges until a single region remains.
The result is a hierarchy of nested regions in which every multi-voxel region splits into exactly two subregions (see the first panel of Fig~\ref{fig:summary}).

For a given number of regions, a natural segmentation objective minimizes the mean squared error summed across their GLMs, which agglomerative clustering pursues greedily, one merge at a time:
\begin{equation}
	\arg \min \sum_r \Tr{E_r}
	\label{eqn:mse_segment_obj}
\end{equation}

We quantify $E_r$ by its trace rather than its determinant\footnote{Computational need aside, minimizing the sum of determinants $|E_r|$ would have been preferable, as this (greedily) maximizes the likelihood of the GLM models (see Appendix \ref{sec:normal_identity}).  The $b=1$ case offers some consolation: trace and determinant agree there.}, which keeps the objective consistent with Ward's method~\parencite{ward1963hierarchical} and inherits its efficient implementation~\parencite{murtagh2014ward,scikit-learn}.
Ward's method minimizes the Within-Cluster-Sum-of-Squares (WCSS), summed over all clusters:
\begin{align*}
	\arg \min \sum_r WCSS(r) & =
	\arg \min \sum_r \sum_{z \in r} \Tr{(z - \mu_r)(z - \mu_r)^T}
	\\
	                         & = \arg \max \sum_r |r| ||\mu_r||^2 \numberthis \label{eq:wards_obj}
\end{align*}
where $z$ are vector-valued observations from cluster $r$, whose mean is $\mu_r$.
Our objective (\ref{eqn:mse_segment_obj}) is nearly identical:
\begin{align*}
	\arg \min \sum_r \Tr{E_r} & =
	\arg \min \sum_r \Tr{Y_r'^T Y_r' - Y_r'^T Q' Q'^T Y_r'}
	\\
	                          & = \arg \max \sum_r |r| || Q^T \bar{Y}_r||_F^2
	\label{eqn:mse_segment_obj_detail} \numberthis
\end{align*}
where we use (\ref{eqn:e_multi_voxel_first_line}) in the first line and the second line uses (\ref{eqn:q_r_multi_voxel}) and observes that $\sum_r \Tr{Y_r'^T Y_r'}$ is constant across all voxel-to-region assignments.
GLOW's segmentation is therefore Ward's method run on $Q^T Y_v$, each voxel's imaging features multiplied by $Q^T$ and vectorized.
In effect GLOW throws away each voxel's residual component before clustering, which makes sense: that component is GLM error whatever happens, so there is nothing to gain by making it consistent across a region.

The same argument raises a second question: should we discard the components of $Y_r$ lying in the span of the nuisance variables $Q_0$ as well?
Those directions are projected out of both $E$ and $H$, so LLR is already invariant to shifts along them.
Perhaps we would do better to \textbf{Focus} on the components of $Y_r$ which the variables of interest $Q_1$ can explain.
We test each objective empirically in Section~\ref{ssec:result_segmentation}; GLOW uses Focus.
\begin{table}[htbp]
	\centering
	\begin{tabular}{lll}
		\toprule
		Method    & Objective                                           & Removes components of ...   \\
		\midrule
		Naive     & $\displaystyle\sum_r |r|\, \|\bar{Y}_r\|_F^2$       & none                        \\
		GLM Error & $\displaystyle\sum_r |r|\, \|Q^T \bar{Y}_r\|_F^2$   & GLM residuals               \\
		Focus     & $\displaystyle\sum_r |r|\, \|Q_1^T \bar{Y}_r\|_F^2$ & GLM residuals \& Covariates \\
		\bottomrule
	\end{tabular}
	\caption{Segmentation objectives: three varieties of Ward's method.}
	\label{tab:seg_objectives}
\end{table}

One observation before moving on: the spatial covariance $S$ inside $E$ (\ref{eqn:e_multi_voxel}) lends the objective a sense of anatomical relevance.
Penalizing spatial covariance discourages merges which conflate distinct tissues, as distinct tissues often carry distinct imaging features.
All three objectives apply some such penalty and differ only in the directions of $\mathbb{R}^N$ along which they measure it, each being Ward's method on a projected voxel ($Y_v$, $Q^T Y_v$ or $Q_1^T Y_v$).
Naive measures spatial covariance in all $N$ directions, GLM Error only in the $a$ directions of $X$, and Focus in the $a - a'$ directions of interest alone (orthogonalized to the covariates).

That description is accurate, but in practice Focus all but ignores spatial covariance when building regions.
Analyses conventionally append a bias column of ones to $X$ so that the GLM is robust to feature offsets \parencite{bishop2006pattern}.
The bias is usually a covariate, not a variable of interest: treating it as one tests the typically nonsensical hypothesis that some imaging feature is non-zero when averaged across the $N$ images.
The bias column therefore lies in the span of $Q_0$, and since $Q_1$ is orthogonal to $Q_0$, multiplying by $Q_1^T$ strips each voxel's across-image mean.
Focus is then measuring spatial covariance among voxels which have each been centered on their own mean, and little covariance survives that centering.
Much of the anatomical relevance which motivated the penalty is discarded along with those means.

The objectives themselves motivate some pre-processing.
Each sums squares over the $b$ imaging features on their native scales, so whichever feature happens to carry the largest variance dominates the merges while the others go overlooked.
The objectives also ignore correlations between features, which lets variance shared by two of them be counted twice.
We therefore pre-process all $Y_v$ by de-meaning, normalizing each feature's scale, and then applying Principal Component Analysis (PCA).
Every voxel of every image goes into the mean and covariance estimates used for pre-processing.

\subsection{Significance Testing}\label{ssec:method:sig_test}
GLOW's inference follows the framework of \textcite{winkler2014permutation}, reusing that machinery (Freedman--Lane residual permutation and Westfall-Young max-statistic FWER) essentially unchanged.
We review it before designing around the two GLOW-specific challenges which arise.

Permutation testing manipulates the observed data to produce statistics consistent with the null hypothesis.
Specifically, we adopt the \textit{Freedman--Lane} \parencite{freedman1983nonstochastic} approach, in which residuals from a reduced model are shuffled among observations.
That is, each voxel's $Y_v$ is replaced with the permuted version:
\begin{equation}
	Y_v^* = P (I_N - Q_0 Q_0^T) Y_v + Q_0 Q_0^T Y_v,
	\label{eqn:freedman_lane}
\end{equation}
where $P$ is a permutation matrix obtained by row-shuffling the identity matrix $I_N$.
When testing a single hypothesis (i.e.\ region), one rejects the null when the original, unpermuted statistic exceeds\footnote{Assuming that larger statistics are less consistent with the null hypothesis, which is true for all statistics mentioned here except Wilks' Lambda.} the $(1 - \alpha)$ percentile of the permuted statistics, a reference set which includes the unpermuted statistic itself \parencite{Phipson2010Permutation}.
This ensures the probability of a Type I error is no more than $\alpha$.

Our images are only block-exchangeable \parencite[see][]{winkler2014permutation}: every voxel's residual from image $s$ must be re-assigned together, to one image $s'$.
GLOW therefore applies a single $P$ across all voxels within a permutation.

To control the FWER while maintaining statistical power, we apply the \textit{Westfall-Young} permutation procedure \parencite{westfall1993resampling}.
Each permutation records only the maximum statistic taken across all hypotheses (i.e.\ regions).
A region is declared significant when its unpermuted statistic exceeds the $1 - \alpha$ percentile of those maxima.
This holds the FWER at or below $\alpha$ under the global null, where no region carries an effect (weak control).
Control in the strong sense, over the null regions of an image which also holds an effect, further requires subset pivotality \parencite{westfall1993resampling}: the null regions' statistics must be distributed the same whether or not the effect is present.
GLOW builds its family of regions from the data, so the effect shapes that family and nothing guarantees the condition; Section~\ref{ssec:result_null} finds it violated on white noise at strong effects.
Testing a full hierarchy of nested regions in this way echoes hierarchical testing \parencite{meinshausen2008}, which localizes an effect by descending a tree of correlated variable clusters under a size-weighted Bonferroni rule; GLOW instead optimizes its hierarchy for GLM fit and controls FWER with a simultaneous max-statistic over all regions rather than a top-down descent.

Two issues arise when applying Westfall-Young in GLOW's context.
First, GLOW tailors its regions to the effects present in the unpermuted images.
Reusing those same regions on the permuted image sets, for which they were never optimized, would hand the observed statistics an advantage the null draws never had.
Selecting regions on the observed data and then testing them there is circular \parencite{kriegeskorte2009circular}, and it breaks the exchangeability on which the FWER bound rests \parencite{hemerik2018exact}.
GLOW restores exchangeability by building a separate hierarchical segmentation for each permuted image set, then collecting the maximum statistic of each hierarchy into the comparison set.

The second issue concerns how each region's statistic is prepared before the regions compete for the maximum: raw statistics are not comparable across regions of different sizes.
Consider the Focus form of (\ref{eqn:mse_segment_obj_detail}), which by (\ref{eqn:h_multi_vox}) is just $\sum_r \Tr{H_r}$.
GLOW builds each region to maximize the $\LLR$'s numerator, but the optimization does not benefit all region sizes equally.
Single-voxel and whole-image regions, for example, are the same in every hierarchy, so their statistics gain nothing from it.
Without adjustment, a maximum over raw statistics would be dominated by the region sizes with the most flexibility to fit the data (often, regions of nearly the whole image).
We need a way to remove this bias.

\begin{figure}[htbp]
	\centering
	\includegraphics[width=\textwidth]{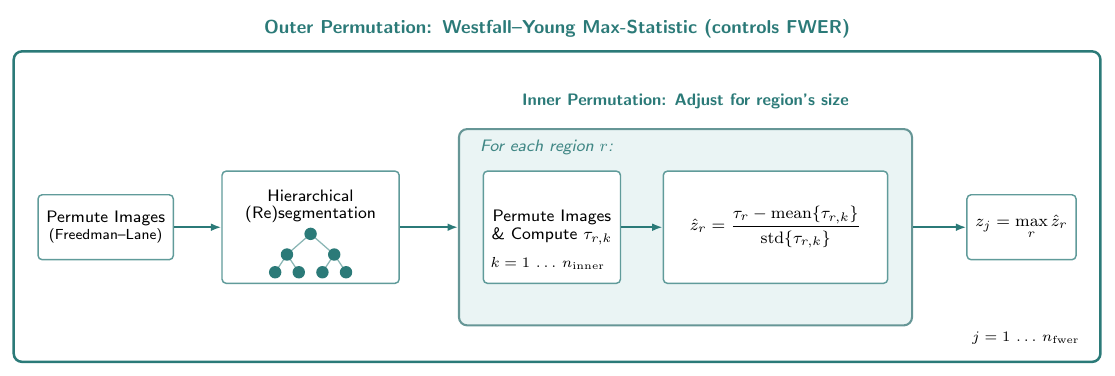}
	\caption{GLOW's nested permutation.
		Each outer Westfall-Young draw ($j = 1 \ldots n_{\mathrm{fwer}}$) re-segments the permuted images and records $z_j = \max_r \hat{z}_r$.
		Calibrating a region (shaded) nests an inner loop ($k = 1 \ldots n_{\mathrm{inner}}$) that recomputes the region's statistic as $\tau_{r,k}$ to form its null $\{\tau_{r,k}\}$, then z-scores the observed $\tau_r$ against it, making $\hat{z}_r$ comparable across region sizes.}
	\label{fig:nested_perm}
\end{figure}

We obtain it with a second, nested\footnote{Please see Section~\ref{ssec:result_computation} for discussion of GLOW's increasingly urgent computational needs.} permutation loop (Fig~\ref{fig:nested_perm}).
Write $\tau_r$ for region $r$'s observed statistic (GLOW uses its $\LLR$, though any of the MANCOVA statistics would serve).
Recomputing it under $n_{\mathrm{inner}}$ permutations of a \emph{fixed} hierarchy yields the null draws $\{\tau_{r,k}\}_{k=1}^{n_{\mathrm{inner}}}$.
Counting the observed statistic among them as $\tau_{r,0} = \tau_r$, as a permutation test counts the identity permutation, we calibrate by z-scoring,
\begin{equation}
	\hat{z}_r = \frac{\tau_r - \operatorname{mean}\{\tau_{r,k}\}_{k=0}^{n_{\mathrm{inner}}}}{\operatorname{std}\{\tau_{r,k}\}_{k=0}^{n_{\mathrm{inner}}}},
	\label{eqn:zscore}
\end{equation}
with the sample standard deviation.
Each $\tau_r$ is compared against references drawn for its own region, so the inflation that region's size and shape carry is accounted for.
This inner calibration nests inside the outer Westfall-Young loop of the first issue: each outer permutation $j$ re-segments the permuted images, calibrates every region of the resulting hierarchy through its own inner loop, and contributes that hierarchy's maximum $z_j = \max_r \hat{z}_r$ to the FWER null.
A region is significant when its observed $\hat{z}_r$ exceeds the $1-\alpha$ percentile of these per-permutation maxima $z_j$.

Reassuringly, the weak FWER bound holds regardless of how well the z-scoring equalizes regions\footnote{The inner calibration is applied identically on every outer permutation, including the observed (unpermuted) draw, so the per-permutation maxima remain exchangeable under the global null and the Westfall-Young bound is exact there \parencite{lehmann2005testing,hemerik2018exact}.}.
The calibration only steers power, giving regions of every size a fair chance at the maximum; it cannot inflate the false-positive rate under the global null.

\subsection{Region Pruning} \label{ssec:prune}
If all has gone well, GLOW has flagged a region matching each true effect's extent as significant.
It has almost certainly flagged that region's relatives in the hierarchy too: a strong effect lends its significance to the smaller fragments nested inside it and to the larger regions which contain it.
These overlapping regions are redundant descriptions of one effect, and GLOW's final step prunes them away, leaving output regions which capture all, and only, the voxels of a single effect's extent.

\begin{figure}[htbp]
	\centering
	\includegraphics[width=0.35\textwidth]{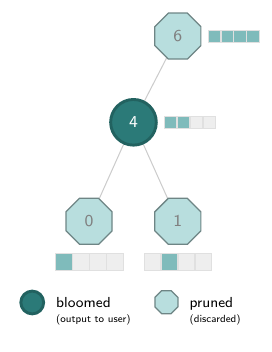}
	\caption{The pruning result for the running example from Fig~\ref{fig:summary}.
		Region~4 is \emph{bloomed} (output to the user).
		The smaller regions nested inside it (0 and 1) and the larger region containing it (6) are \emph{pruned}, their significance explained away by their overlap with region~4.}
	\label{fig:prune_hierarchy}
\end{figure}

Let $B$ denote the set of \emph{bloomed} regions, those output to the user; every other significant region is \emph{pruned} (Fig~\ref{fig:prune_hierarchy}).
A valid $B$ must meet two requirements.
First, its regions must be pairwise disjoint: homogeneous effects cannot meaningfully overlap, so neither should the regions we report.
Second, every significant region must intersect some member of $B$; otherwise we would silently discard a significant result.
Together these make $B$ a tree-cut (similar in spirit to \textcite{michel2012}).

One natural choice of $B$ maximizes the likelihood of the observed images.
Under this model each voxel follows the $\beta_r$ of whichever bloomed region contains it, while a voxel in no member of $B$ follows the reduced model's $\beta_r$, contributing a convenient LLR of zero.
Spatial independence of the residuals makes the objective tractable: the log-likelihood ratio over all significant voxels is just the sum of the per-region ratios, $\sum_{r \in B} \LLR(r)$.
Maximizing this sum has a clean global optimum, reached by a bottom-up Dynamic Programming (DP) solution over the hierarchy \parencite{salembier2000binary,guigues2006scale}.

The principled simplicity of DP is, however, too good to be true.
Partition a region $r$ into disjoint pieces $r_1, \ldots, r_k$; the whole-region LLR never exceeds the sum over those pieces:
\begin{equation}
	\LLR(r) \leq \sum_{i=1}^{k} \LLR(r_i).
	\label{eqn:llr_superadd}
\end{equation}
A single $\beta_r$ cannot follow the differing spatial means of $r_1, \ldots, r_k$, so pooling them adds to $E$ while subtracting from $H$ (see Appendix~\ref{sec:pool_error_hypothesis}).
DP therefore sacrifices completeness: offered a whole region or a partition of it, DP never prefers the whole.
This violates our stated goal of outputting one region per effect; Section~\ref{ssec:result_pruning} confirms it empirically.

An alternative maximizes $\LLR$ greedily: \emph{bloom} the significant region of maximum $\LLR$, \emph{prune} every region which intersects it, and repeat until no candidates remain.
This errs in the opposite direction, sacrificing homogeneity: it favors a large region, which may absorb non-effect volume alongside an effect's extent.
GLOW adopts greedy pruning on the strength of its empirical results (Section~\ref{ssec:result_pruning}).
Forgoing the maximum-likelihood DP for its greedy approximation is uncomfortable\footnote{We spent significant effort trying to redeem the DP with regularization penalties which reward larger regions over their smaller constituents; a geometric prior over $|B|$, for instance, adds a constant penalty $\lambda|B|$ to the objective and coexists gracefully with the dynamic program.}; Section~\ref{sec:discussion} blames the ghost of the invalid independence assumption.

\section{Experiment}\label{sec:experiment}
Evaluating detection performance requires ground truth: to call a significant voxel a true or false positive, and thus to compute sensitivity or PPV, we must know which voxels carry an effect.
Real image sets do not supply it.
To validate in conditions as close to real images as possible, we impose a synthetic effect of known extent onto real data.
Imposing an effect also puts its character under our control: turning the effect's LLR knob up, we watch detection performance change as effects grow from subtle to obvious.

We test on two image sets: White Gaussian Noise (WGN), which satisfies our modeling assumptions (independent and identically distributed Gaussian residuals), and diffusion MRI maps from the Human Connectome Project (HCP), which carry the realistic spatial structure of a brain.
Section~\ref{sec:data_code_availability} provides the HCP diffusion processing details and instructions on how to repeat our experiments.

\begin{figure}[htbp]
	\centering
	\includegraphics[width=\textwidth]{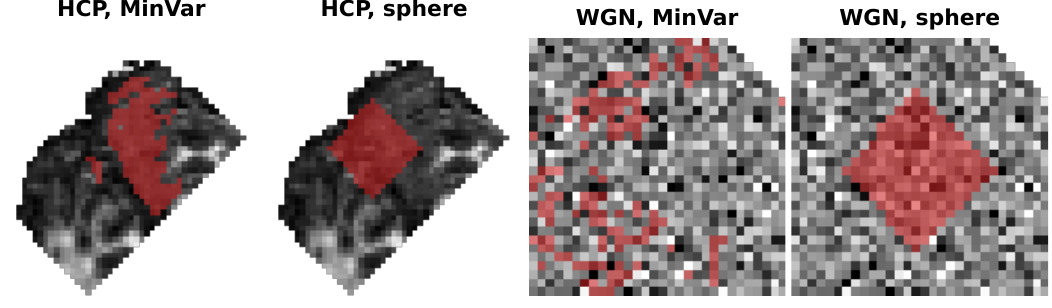}
	\caption{The two extent shapes on each source, each of 2{,}500 voxels, planted on the same images: one seed's extents (red) on a shared axial slice of one of those images (HCP: FA; WGN: noise).}
	\label{fig:plants}
\end{figure}

Each trial aims to plant one effect.
For either image set we first sample the effect's extent, the set of voxels it will occupy.
Starting from a single seed voxel chosen uniformly at random, we grow the extent to a target size, at each step annexing whichever neighboring voxel minimizes the resulting region's spatial variance $\Tr(S)$ (Fig~\ref{fig:plants}).
We call it the MinVar extent.

On HCP the MinVar extent follows tissue, so its voxels share a mean and its spatial variance $S$ stays small; on WGN, which has no tissue to follow, it is a one-voxel-thick dendrite spread through the crop.
We also plant a shape which ignores the images: a ball of the same size centered in the crop, grown by repeated face-connected dilation (an octahedron, the ball of the voxel grid's own metric).
Each sphere is planted on the same images as its MinVar twin.
On HCP a sphere typically straddles tissue boundaries, so its voxel means differ and $S$ is large, and the offset which brings the whole region to its target LLR must be larger than a MinVar extent of the same size would need.
Such a sphere is arguably not one effect but several: the voxels on either side of an anatomical boundary share the planted offset but not a common $\bar{Y}$, and so not one GLM in the sense of Section~\ref{ssec:problem}.

\begin{figure}[htbp]
	\centering
	\includegraphics[width=\textwidth]{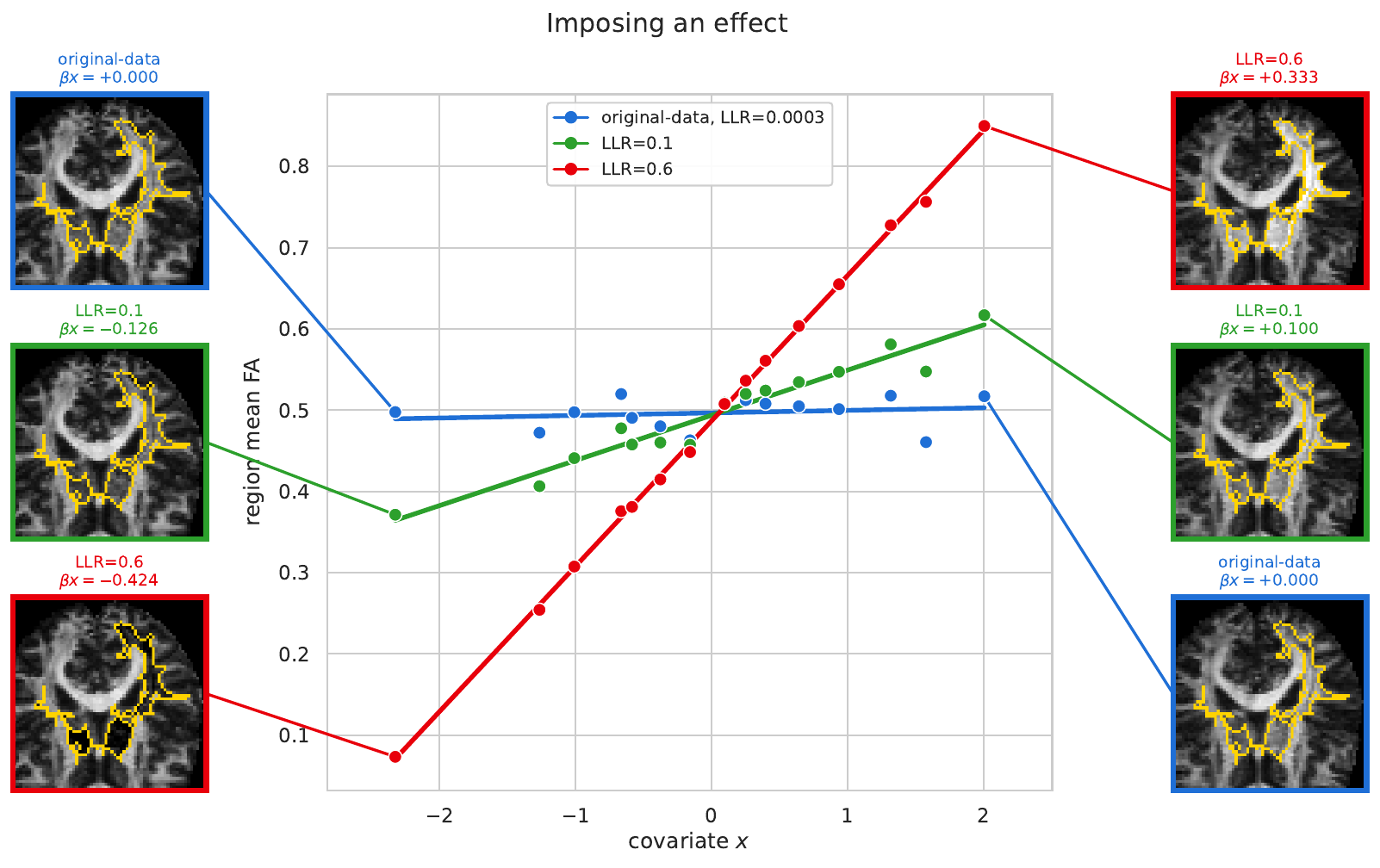}
	\caption{Imposing a synthetic effect.  Green and red mark moderate and severe effects imposed onto the original image set, in blue.
		The left column gives one image of the original set alongside its modified versions at each effect strength, each labeled with the FA offset the effect induces, $\beta x$.
		The right column repeats this for another image, the one with maximum $x$.}
	\label{fig:impose_effect}
\end{figure}

Within either extent the real images supply the starting data $Y'$.
We draw the regressors of interest at random from a standard normal, so any initial correlation between the images and $X$ is purely coincidental, and append a bias column of ones as the sole covariate.
We then add a spatially constant offset $\Delta \in \mathbb{R}^{N \times b}$, the same in every voxel of the extent, chosen to bring the region to a target LLR (Fig~\ref{fig:impose_effect}).
Many offsets reach that target; we take the one with the smallest mean-squared change from the original images, keeping the perturbed image set as faithful to the original as we can.
Appendix~\ref{sec:impose_constant_offset} gives the optimization detail.


\section{Results}\label{sec:results}
Unless otherwise noted, all experiments use the following defaults.
Each source (WGN or HCP) is cropped to a single connected sphere of $25{,}000$ voxels (about a tenth of a full-brain volume at HCP resolution), with $N = 100$ subjects.
Each trial uses a single imaging feature ($b = 1$), on HCP sampled from our six microstructural maps (FA, MD, MK, ICVF, ISOVF, and OD; see Section~\ref{sec:data_code_availability}); the detection experiments also run at $b = 2$, drawing two of the maps per seed.
Synthetic effects take the MinVar extent unless a sphere is named.
The design matrix holds one regressor of interest and the bias column ($a = 2$, $a' = 1$).
Synthetic effects span $10\%$ of the crop.
We set effect strength with LLR-per-voxel, which is closer to $|r|$-invariant than raw LLR, sweeping eleven logarithmically spaced values from $0.003$ to $0.3$.
Experiments which do not sweep effect strength use that grid's midpoint, $0.03$.
Both sources sit on a $2$~mm isotropic grid, so a kernel width in millimeters means the same on each.
The voxel-wise baselines smooth every image with a $2$~mm FWHM Gaussian kernel (Appendix~\ref{sec:stat_selection}); GLOW does not smooth.
GLOW uses $n_{\mathrm{inner}} = 250$ inner permutations.
Each configuration runs 50 random seeds.
Significance testing for all methods bounds FWER at $\alpha = 0.05$ using $n_{\mathrm{fwer}} = 500$ outer (Westfall-Young) permutations.
That is fewer permutations than a single published study ought to use, but it buys us comprehensive testing across many contexts here.

All performance comparisons report Dice, sensitivity, and positive predictive value (precision) against the planted ground-truth extent, together with the homogeneity and completeness of Section~\ref{ssec:problem}.
We report PPV rather than specificity because effects cover only a small fraction of the volume: specificity is then dominated by the large true-negative pool and stays near one even for a method which flags many false positives~\parencite{maierhein2024metrics}.

With a single planted effect, homogeneity and completeness each reduce to a simpler question.
Completeness asks whether the detection comes back as one region, and is nearly binary.
A trial that returns one region leaves no uncertainty in $R$ for $T$ to resolve, so $H(R) = 0$ and the convention scores it $1$.
A trial that returns several regions scores near $0$, however cleanly those regions tile the effect: both scores see detected voxels alone, so as detection gets cleaner nearly every detected voxel carries the same label $T = 1$, and $H(T)$ vanishes, taking the numerator $I(R, T) \leq H(T)$ with it.
Mean completeness is therefore close to the proportion of trials which return a single region.
Homogeneity asks whether the region boundaries separate effect voxels from false positives.
It credits only what the split into regions tells a reader about $T$ beyond the detected volume taken as a whole, so a single mixed region scores $0$ however much of it is effect, while many small pure regions score near $1$.
Put plainly, completeness rewards handing the detection back in one piece, homogeneity rewards pieces that are each all effect or all false positive, and neither counts how much of the effect was found, which Dice and sensitivity measure.

\subsection{Tuning: Segmentation Objective}\label{ssec:result_segmentation}
\begin{figure}[htbp]
	\centering
	\includegraphics[width=\textwidth]{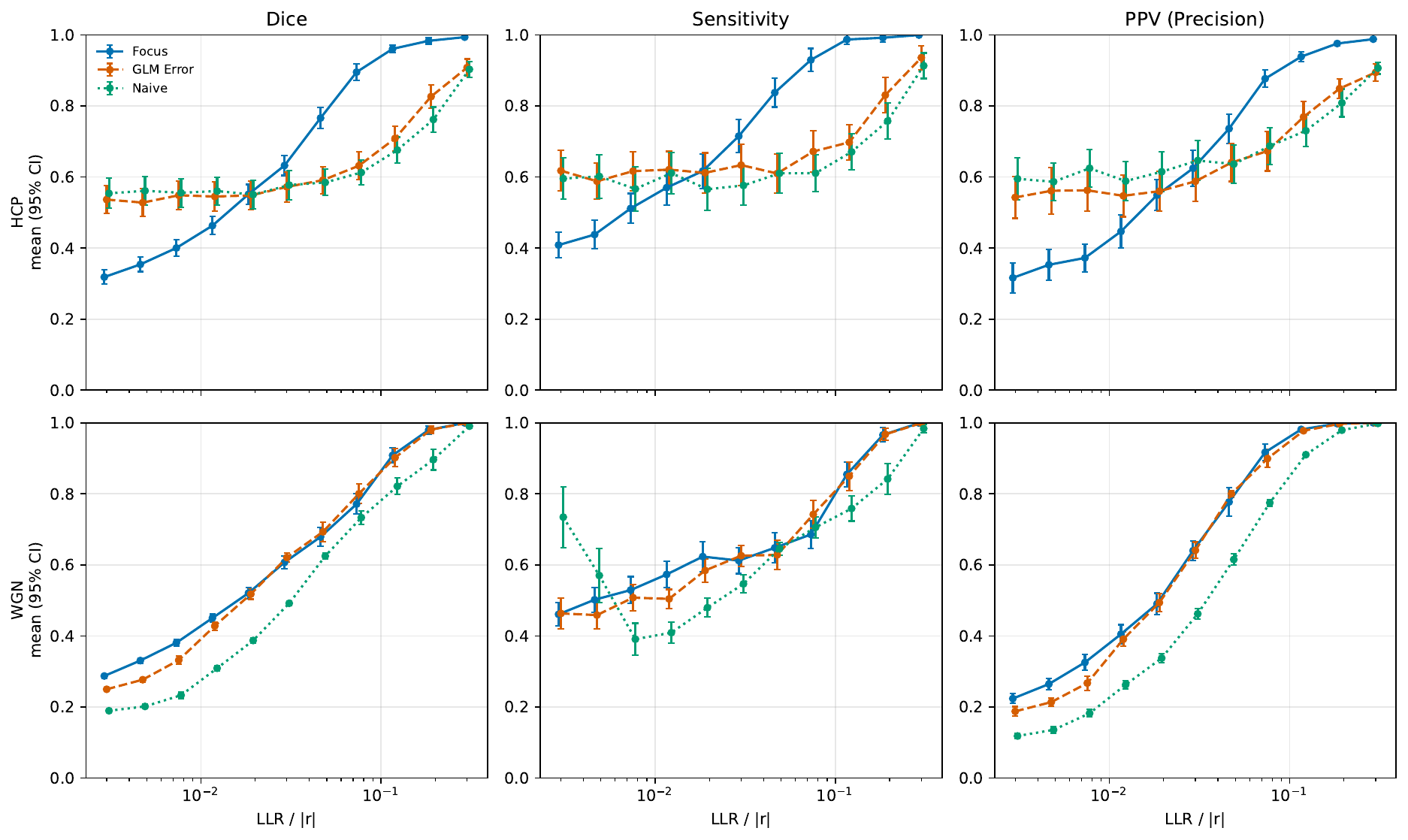}
	\caption{Segmentation quality. Oracle Dice of the best-matching region in the hierarchy for each of the three Ward's objectives (Table~\ref{tab:seg_objectives}).  Markers show the mean across random seeds; error bars give its $95\%$ confidence interval ($\pm 1.96$ standard errors of the mean).}
	\label{fig:segment}
\end{figure}

We compare the three Ward's objectives of Table~\ref{tab:seg_objectives}.
Ideally a segmentation produces some region which matches the effect's extent, so we score it by oracle Dice: the largest Dice overlap with the ground truth extent taken across every region of the hierarchy (Fig~\ref{fig:segment}).
Of course, oracle Dice cannot be computed in practice; we use it here to isolate segmentation quality.

Comfortingly, every objective's oracle Dice climbs toward one as the effect strengthens.

On WGN data, discarding the GLM residuals before segmentation (as both GLM Error and Focus do) consistently improves Dice.

On realistic images (HCP), Focus performs admirably at high LLR, suggesting that the reduced model's GLM error may indeed be a distraction.
We suspect its relative underperformance at low LLR is an artifact of how we sample effect extents: they are grown to minimize spatial variance, which is exactly what Naive Ward's minimizes.
GLM Error, whose $Q$ retains that critical bias column $\vec{1}$ in the span of $Q_0$, inherits much of the same advantage.
Focus is left adrift here because it strips each voxel's across-image mean (see the Focus discussion of Section~\ref{ssec:method:hier_seg}).

The GLM Error and Naive objectives are, we would argue, ``cheating'' here: they reproduce the very mechanism which sampled the effect extent.
GLOW therefore uses Focus, on the strength of its high-LLR performance and the fairness concern with the alternatives.

Neither Focus's strength at high LLR nor its weakness at low LLR appears on WGN data.
There a much higher proportion of the GLM error is spatial variance $S$ (\ref{eqn:e_multi_voxel}), so there is less to gain by focusing on the $Q_1$ component of $\bar{Y}_r$.

\subsection{Tuning: Region Pruning}\label{ssec:result_pruning}
\begin{figure}[htbp]
	\centering
	\includegraphics[width=\textwidth]{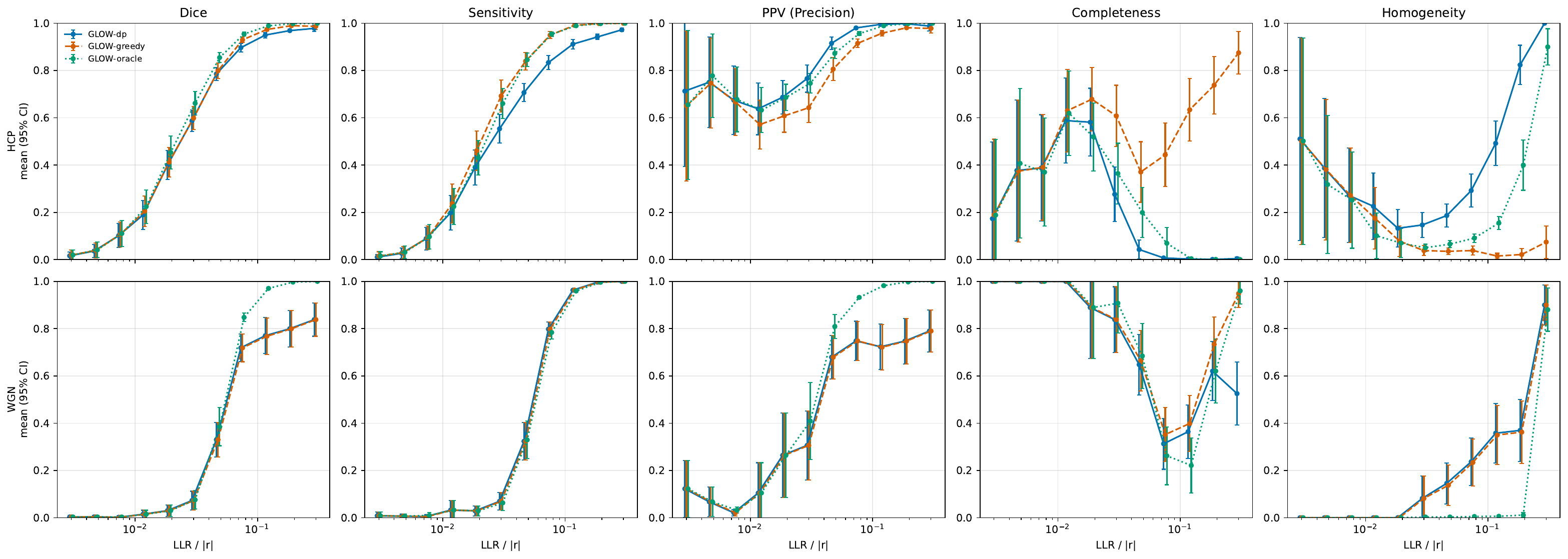}
	\caption{Pruning rule comparison, under the Focus segmentation objective.  Markers show the mean across random seeds; error bars give its $95\%$ confidence interval.}
	\label{fig:prune}
\end{figure}

We next compare the two pruning rules of Section~\ref{ssec:prune}: greedily blooming the maximum-LLR significant region, or using Dynamic Programming (DP) to find the disjoint set of significant regions which maximizes likelihood.
Segmentation and hypothesis testing cap the Dice any pruning rule can reach, as sometimes no strong candidate sits among the regions declared significant.
To quantify that ceiling we add an ``Oracle'' rule, which selects the $B$ of maximum Dice.
The rules share one input: each experiment (a seed, effect strength and source, WGN or HCP) is segmented and tested once, and every rule prunes from that same set of significant regions (Fig~\ref{fig:prune}).

On Dice the two rules are hard to separate.
HCP's average is $0.551$ for greedy against $0.537$ for DP, with DP winning only $15\%$ of seed-paired comparisons.
On WGN they tie ($0.326$ each).
On whether an output region holds all (completeness) and only (homogeneity) of one effect, the difference is stark.
As effect strength rises on HCP, greedy does a fine job of collecting whole effects (completeness $\rightarrow 1$) and DP does just the opposite.
As Section~\ref{ssec:prune} anticipated, partitioning a region into smaller outputs can only raise the LLR which DP maximizes, so completeness suffers.
The flaw runs deep: at strong enough effects every individual voxel of an extent is declared significant, and (\ref{eqn:llr_superadd}) leaves DP no reason to bundle them back together.
DP's homogeneity is certainly higher, though that counts for little in this single-effect setting.
At strong effects, where Dice approaches one, almost every output voxel\footnote{Recall that homogeneity and completeness are computed only over voxels output to the user.} belongs to the one effect, leaving little uncertainty in $T$ for $R$ to resolve.

GLOW is therefore built on greedy pruning and Focus segmentation, sidelining a global optimum in favor of its approximation; Section~\ref{sec:discussion} traces that choice back to the false independence assumption in our LLR.

\subsection{Type I Error Control}\label{ssec:result_null}
\begin{figure}[htbp]
	\centering
	\includegraphics[width=.7\textwidth]{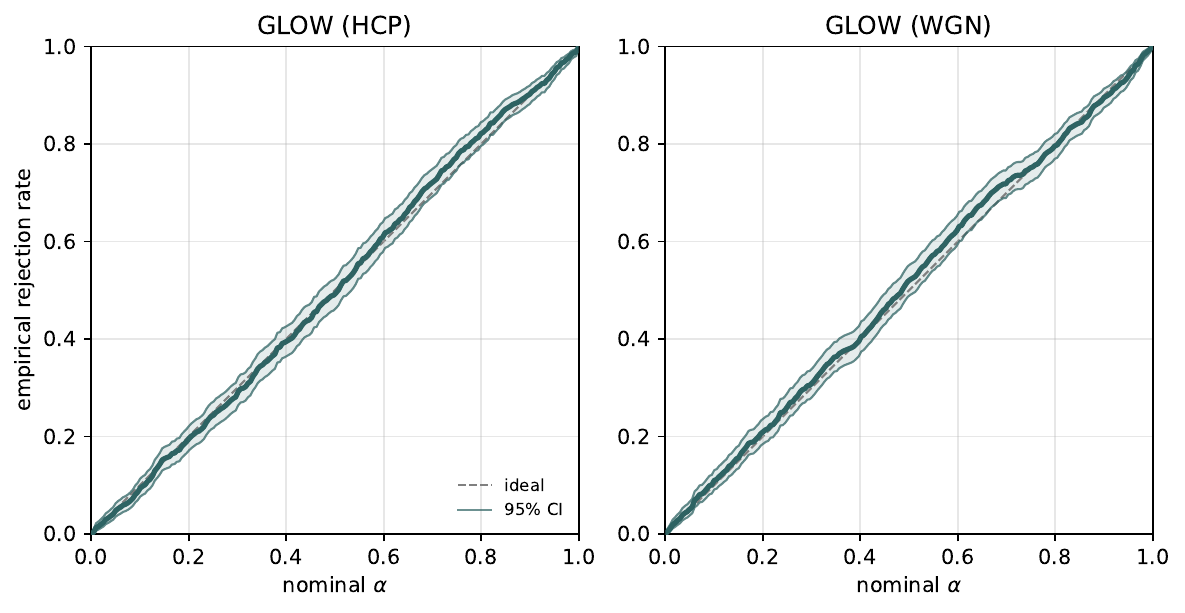}
	\caption{FWER control.  Shading is the pointwise $95\%$ interval for the true rejection rate given the 1{,}000 null seeds, so a calibrated method's band brackets the dashed ideal.}
	\label{fig:null_calibration}
\end{figure}

Before evaluating detection performance, we verify that GLOW controls the FWER.
Under the null hypothesis (no imposed effect) we run 1{,}000 seeds of HCP and WGN images.
We report the percentage of experiments which declared any region significant, the empirical FWER, against the nominal rate (Fig~\ref{fig:null_calibration}).

On both sources the empirical rejection rate tracks the nominal $\alpha$ across its whole range, the $95\%$ band bracketing the ideal line ($4.8\%$ on HCP and $5.1\%$ on WGN at $\alpha = 0.05$), so at these 1{,}000 seeds GLOW's FWER under the global null is indistinguishable from nominal.

That is weak control.
Strong control asks more: in an image which does hold an effect, the chance of rejecting any region containing none of it must still not exceed $\alpha$.
The detection experiments of Section~\ref{ssec:result_detection} measure this directly, since an output region without a single planted voxel is a false rejection of a true null.
Table~\ref{tab:false_region} counts the trials which return one.
On HCP GLOW's count never exceeds 5 of 50, indistinguishable from nominal (one-sided binomial $p \geq 0.10$), though suggestive.
On WGN the count holds at or near zero through the weaker effects, then climbs with strength to as many as 25 of 50, well above $\alpha$.
That climb is the risk anticipated in Section~\ref{ssec:method:sig_test}: the effect itself raises the null regions' chance of reaching significance.
VBA's and CET's counts do not move with strength, their false regions being the same few seeds' null clusters throughout; VBA-TFCE's vary more, and on WGN MinVar at $b = 2$ it too exceeds $\alpha$ at each of the six strongest effects (6 to 9 of 50).

\begin{table}[htbp]
	\centering
	\begin{tabular}{lrrrrrrrr}
		\toprule
		            & \multicolumn{4}{c}{HCP} & \multicolumn{4}{c}{WGN}                                                                 \\
		\cmidrule(lr){2-5} \cmidrule(lr){6-9}
		            & \multicolumn{2}{c}{MinVar} & \multicolumn{2}{c}{Sphere} & \multicolumn{2}{c}{MinVar} & \multicolumn{2}{c}{Sphere} \\
		\cmidrule(lr){2-3} \cmidrule(lr){4-5} \cmidrule(lr){6-7} \cmidrule(lr){8-9}
		LLR$/|r|$   & $b{=}1$ & $b{=}2$ & $b{=}1$ & $b{=}2$ & $b{=}1$     & $b{=}2$     & $b{=}1$     & $b{=}2$     \\
		\midrule
		$0.0189$    & 1       & 0       & 1       & 0       & 0           & 0           & 3           & 0           \\
		$0.0300$    & 1       & 0       & 1       & 1       & 0           & 0           & 2           & 0           \\
		$0.0475$    & 1       & 1       & 1       & 0       & 0           & 0           & 5           & 1           \\
		$0.0754$    & 1       & 0       & 3       & 1       & \textbf{7}  & 2           & 4           & 1           \\
		$0.1194$    & 1       & 0       & 3       & 1       & \textbf{9}  & 4           & 5           & \textbf{6}  \\
		$0.1893$    & 1       & 2       & 5       & 4       & \textbf{10} & \textbf{11} & \textbf{9}  & \textbf{6}  \\
		$0.3000$    & 2       & 2       & 5       & 3       & \textbf{16} & \textbf{25} & \textbf{11} & \textbf{9}  \\
		\bottomrule
	\end{tabular}
	\caption{GLOW trials (of 50) returning a region which holds no planted voxel, a false rejection of a true null.  Bold where the count exceeds $\alpha = 0.05$ (one-sided binomial $p < 0.05$, i.e.\ 6 or more).  The four weakest strengths of the grid ($0.0030$ to $0.0119$) are omitted: every count there is 0 or 1.}
	\label{tab:false_region}
\end{table}
GLOW therefore controls FWER in the weak sense, but on white noise not in the strong sense.
A further region reported alongside an obvious effect deserves less confidence than its significance suggests.

\subsection{Detection Performance}\label{ssec:result_detection}
We compare GLOW against the voxel-wise baselines (VBA, CET, and VBA-TFCE) on detection metrics (Dice, Sensitivity, and PPV) as well as segmentation scores (Completeness and Homogeneity).
Each voxel-wise method uses the statistic and smoothing kernel which jointly optimize its mean Dice on HCP (Appendix~\ref{sec:stat_selection}).
All three smooth at $2$~mm FWHM and use Wilks' Lambda, VBA-TFCE after per-voxel z-scoring; under our single-regressor contrast VBA and CET are invariant to the choice of statistic anyway.
CET uses an initial cluster-forming threshold of $10^{-3}$, as recommended by \textcite{woo2014cluster}; TFCE uses extent and height exponents $e = 0.5$ and $h = 2$, the standard three-dimensional settings \parencite{smith2009threshold}.

\begin{figure}[tp]
	\centering
	\includegraphics[width=\textwidth]{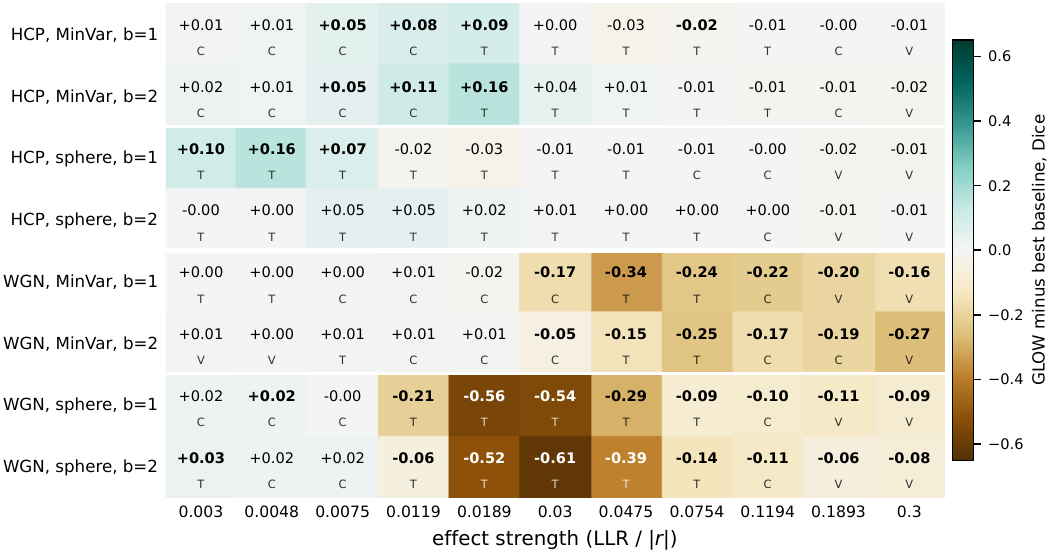}
	\caption{GLOW's Dice minus the best baseline's, by source, extent shape, $b$ and effect strength.  The best baseline is the one with the highest mean Dice in that cell (letter: V for VBA, T for VBA-TFCE, C for CET), and each entry is the mean of the 50 seed-paired differences, bold where its $95\%$ confidence interval excludes zero.}
	\label{fig:gap_map}
\end{figure}

\begin{figure}[tp]
	\centering
	\includegraphics[width=\textwidth]{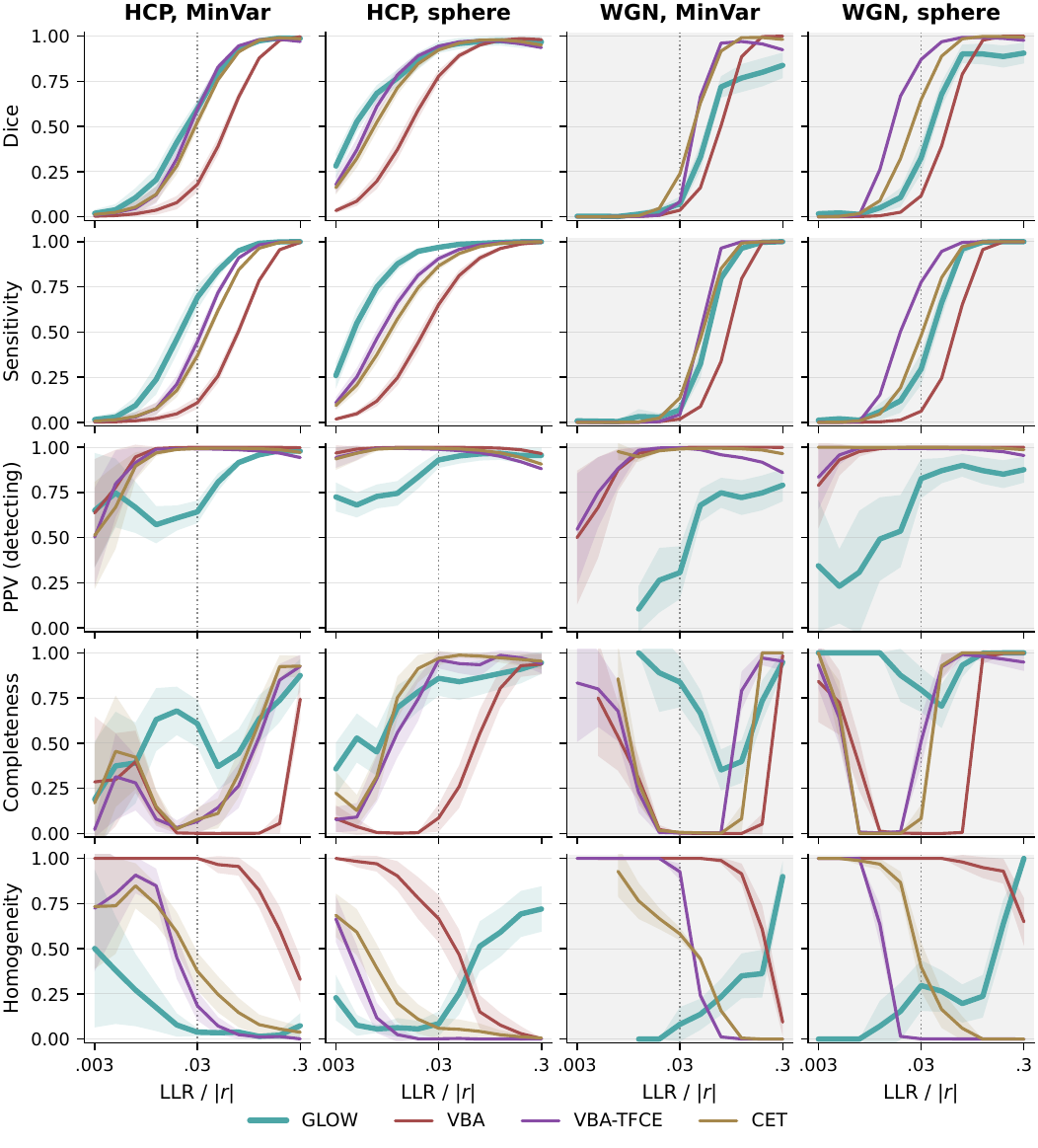}
	\caption{Detection performance versus effect strength ($b = 1$), one metric per row and one source and extent shape per column.  Each curve is the mean across random seeds, shaded with the $95\%$ confidence interval of that mean.  PPV, completeness and homogeneity average over the trials which detect some effect voxel.}
	\label{fig:curves_b1}
\end{figure}

\begin{figure}[tp]
	\centering
	\includegraphics[width=\textwidth]{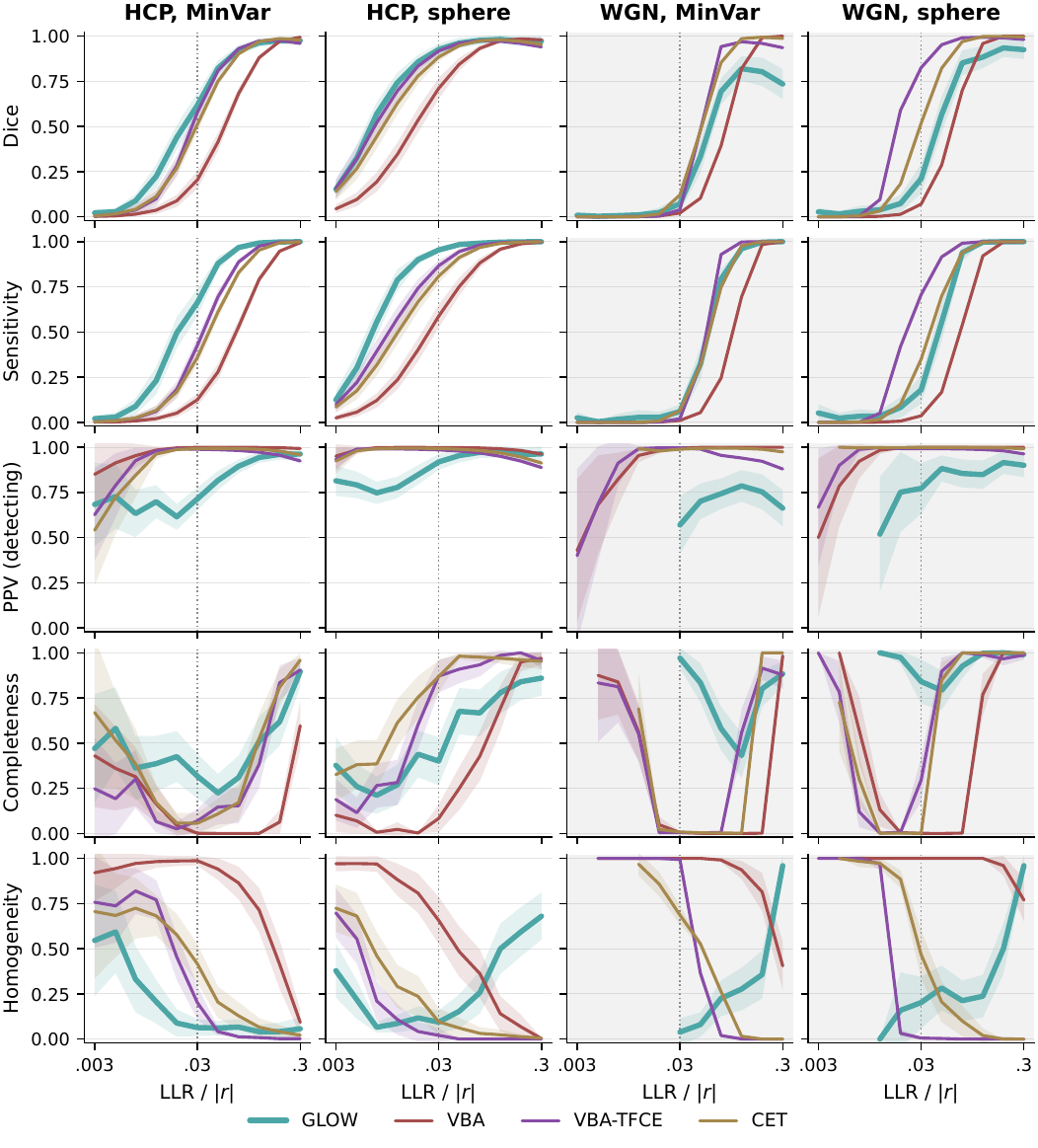}
	\caption{Fig~\ref{fig:curves_b1} at $b = 2$.}
	\label{fig:curves_b2}
\end{figure}

We sweep effect strength in four modes, crossing the two sources with the two extent shapes of Section~\ref{sec:experiment}, at $b = 1$ and $b = 2$.
Fig~\ref{fig:gap_map} summarizes every cell as GLOW's Dice minus that of the baseline with the best mean Dice there, paired by seed; Figs~\ref{fig:curves_b1} and~\ref{fig:curves_b2} give the curves behind it.

The source decides GLOW's standing.
On HCP, GLOW leads on Dice wherever the effect is hard to see and ties once it is easy: averaged over the effect strengths below LLR$/|r| = 0.03$ it gains $0.059$ (MinVar extent, $b = 1$ and $2$), up to $0.16$ in a single cell, and from LLR$/|r| =0.03$ up it stays within $0.04$ of the best baseline in every HCP cell.
On WGN it trails from moderate strengths up, by as much as $0.34$ on the MinVar extent and $0.61$ on the sphere; at strong effects its Dice plateaus while the best baselines converge to $1.00$.
At a given nominal LLR$/|r|$ the HCP sphere is easier to detect than the other three cells.
A sphere straddling tissue boundaries has voxels whose means differ; fit with a common mean, that spread enters the error as the spatial variance $S$ (\ref{eqn:e_multi_voxel}), so reaching a given LLR$/|r|$ takes a larger offset than a MinVar extent would need.
A voxel-wise test fits each voxel its own mean, so it sees the larger offset without the spread: recomputed with each voxel centered on its own mean, the same region's LLR$/|r|$ is a median $2.7$ times its nominal value across the HCP sphere's trials at $b = 1$ ($2.3$ at $b = 2$), against $1.3$ for the MinVar extent on HCP and $1.0$ on WGN, where voxel means do not differ.

On WGN neither shape meets an anatomical boundary, so both carry the same per-voxel strength.
What still separates them is their geometry, which matters to the baselines because they pool signal over neighboring voxels: all three smooth the images, and CET and TFCE also pool statistics over clusters.
A compact sphere gains from that pooling, while a one-voxel-thick dendrite (the MinVar extent on WGN) loses by it.
At LLR$/|r| = 0.0189$, VBA-TFCE's Dice is $0.67$ on the sphere against $0.01$ on the dendrite, while GLOW's rises only from $0.03$ to $0.11$.

Completeness separates GLOW from the baselines on the MinVar extent.
Among trials that detect anything, GLOW returns the effect as one region on $57\%$ of HCP trials against the best baseline's $35\%$ ($45$ against $40\%$ on WGN), and through the HCP mid-range of effect strengths (LLR$/|r|$ from $0.0189$ to $0.0754$) it beats the best baseline's completeness in $70\%$ of seed-paired comparisons.
Completeness here is close to the share of trials which return a single region (across all cells the two correlate at $0.990$).
This is the property Section~\ref{ssec:problem} asks for and the one Dice cannot see: on such an effect the baselines recover comparable volume but hand it back in pieces rather than as a whole region.
On the HCP sphere, which Section~\ref{sec:experiment} argues is several effects rather than one, the gap closes: CET returns it as one region as often as GLOW at $b = 1$ ($70\%$ each) and more often at $b = 2$ ($70$ against $49\%$).

Wholeness costs precision.
GLOW is the more aggressive detector on HCP: its sensitivity beats the best alternative by up to $0.24$ on the MinVar extent and $0.29$ on the sphere, and its PPV is lower at all eleven effect strengths of both, on the MinVar extent winning only $9$ of the $402$ seed-paired comparisons in which both GLOW and a baseline detect something.

GLOW's PPV loss on HCP is largely undersegmentation: a region captures an effect and then keeps growing, bundling non-effect volume along with it.
The alternative explanation, that these false positives make up wholly spurious regions, barely applies on HCP.
Averaged over GLOW's false positive voxels, the region a false voxel sits in holds $72\%$ of the planted MinVar extent ($64\%$ of the sphere), and only $12$ of the $1{,}793$ regions GLOW returned on it hold none of the effect.
This is exactly the error Section~\ref{ssec:prune} predicted of greedy pruning.

It is also why homogeneity runs the other way: on HCP, VBA holds above $0.95$ until the effect is strong enough for its smoothed pieces to merge, while GLOW falls from $0.50$ to $0.04$ by the middle strength effect.
Through the HCP mid-range, $99.7\%$ of VBA's regions lie wholly inside the planted extent against $17\%$ of GLOW's, at a median of $40$ regions against GLOW's $2$.
The baselines return many pure pieces of one effect; GLOW returns few pieces that straddle its boundary.

On WGN the deficit lies in the test rather than the tree.
The best single node of GLOW's own Focus hierarchy matches the MinVar extent about as well on WGN as on HCP (oracle Dice $0.52$ against $0.55$ at LLR$/|r| = 0.0189$; Fig~\ref{fig:segment}), yet GLOW declares any region significant in $9$ of 50 WGN trials there against $44$ of 50 on HCP, at a median smallest $p$-value of $0.18$ against $0.004$.
GLOW's segmentation, running on both the original and the permuted images, identifies the region whose GLM candidate is most compelling.
On HCP data, where the images are smooth, a region's neighbors offer limited variety from which to assemble a consistent GLM under the null hypothesis.
WGN, on the other hand, has ample choice to construct a strong statistic under the null hypothesis, more so on the MinVar extent, whose dendritic tour of the volume offers more neighbors to choose from.
GLOW catches up only once the effect is strong, and then by returning a region larger than the effect, so its WGN PPV stays at $0.66$ to $0.92$ from LLR$/|r| = 0.075$ up.

\begin{figure}[htbp]
	\centering
	\includegraphics[width=.7\textwidth]{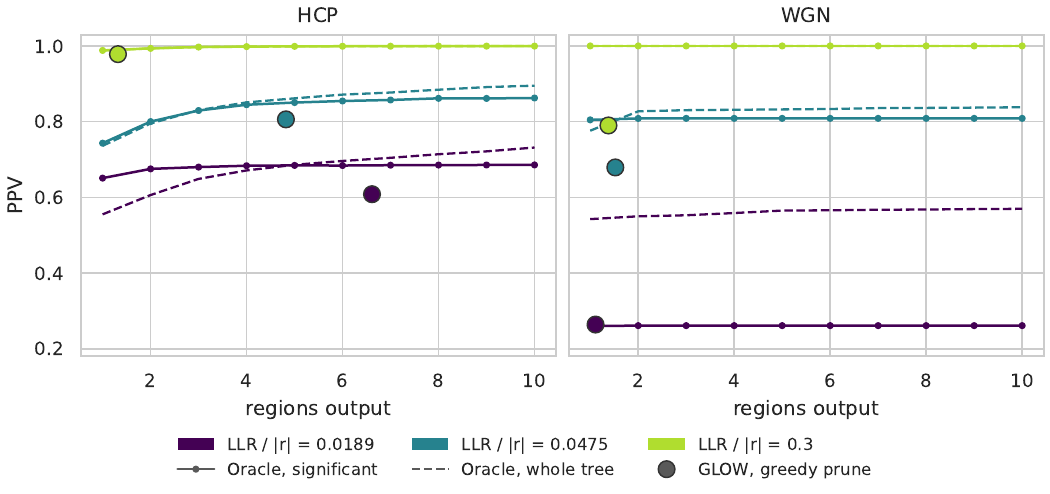}
	\caption{The PPV available to a pruning rule, and to the tree it prunes, for the MinVar extent at $b = 1$, over the same seeds as Fig~\ref{fig:curves_b1}.
		Each line gives the PPV of the union of at most $k$ regions, chosen with knowledge of the ground truth to maximize Dice with the planted extent.
		Solid lines choose from the FWER-significant regions alone, the pool GLOW's greedy rule ranks; dashed lines choose from every node of the Ward tree.
		Circles mark GLOW's own operating point, its mean region count against its mean PPV.
		All three are averaged over the trials in which GLOW discovered something.
		Both Oracles maximize Dice rather than PPV, so the wider pool is not guaranteed to be the more precise one.}
	\label{fig:oracle_vs_k}
\end{figure}

On HCP, by contrast, the cause of the undersegmentation (low PPV) is mostly upstream of the pruning rule.
Fig~\ref{fig:oracle_vs_k} (left) plots PPV against a budget of $k$ output regions for two Oracles, each choosing, with knowledge of the ground truth, at most $k$ regions whose union maximizes Dice against the planted effect.
The first draws only from the FWER-significant regions, the same pool greedy ranks, so no pruning rule could beat it at the same region count; the second draws from any node of the Ward tree.
Stepping from GLOW's output to the first Oracle therefore measures what the pruning rule leaves behind; from the first Oracle to the second, what the significance test withheld; and from the second up to VBA's PPV, what the segmentation never offered.
At weak effects (LLR$/|r| = 0.0189$, MinVar extent) greedy returns a PPV of $0.608$ from a mean of $6.6$ regions (median $1$); with a budget of $k = 10$ regions the first Oracle reaches $0.686$, the second $0.732$, and VBA reaches $0.998$.
Of the $0.390$ separating greedy from VBA, greedy pruning accounts for $20\%$, the significance test for a further $12\%$, and the segmentation for the remaining $68\%$.
In other words, at weak effects pruning can only do so much when segmentation does not isolate the effect's extent somewhere in the tree.

\subsection{Computation}\label{ssec:result_computation}
As described thus far, GLOW is sorely in need of computational optimization.
The central challenge is computing log-likelihood ratios for a hierarchical segmentation quickly.
A region's LLR needs its $E$ and $H$ matrices (\ref{eqn:e_multi_voxel}, \ref{eqn:h_multi_vox}), which depend only on $|r|$, $Y'^T Y'$ and $\bar{Y}_r$.
Given these sufficient statistics for two regions, those of their union follow immediately.
We may therefore walk the tree in topological order, from single voxels up to the whole image, at a per-region cost which no longer grows with region size $|r|$.
Better yet, re-indexing the voxels in the tree's depth-first order leaves every region spanning a contiguous range of voxel indices.
Each of $|r|$, $Y'^T Y'$ and $\bar{Y}_r$ is then the difference between two entries of a single cumulative sum over the voxels, the standard prefix-sum trick.

Even so, the inner calibration of Section~\ref{ssec:method:sig_test} multiplies this workload: every region of every outer (FWER) permutation needs its own $n_{\mathrm{inner}}$ inner permutations to estimate the mean and standard deviation that studentize its LLR (\ref{eqn:zscore}).
A fit therefore pays for $n_{\mathrm{fwer}}$ tree builds, each carrying $n_{\mathrm{inner}} + 1$ LLR computations per region.
This multiplier, not the clustering, separates GLOW's wall time from a voxel-wise test's.

To characterize compute time, we timed each method's fit on HCP images at analysis-crop sizes ranging from 1{,}000 to 224{,}619 voxels (the full connected HCP support), with a single imaging feature ($b = 1$).
A moderate effect (per-voxel LLR of $0.03$ over 10\% of the crop) was imposed so the pruning step had a genuine effect to process.
Each fit ran $n_{\mathrm{fwer}} = 500$ outer (Westfall-Young) permutations, GLOW additionally drawing $n_{\mathrm{inner}} = 250$ inner permutations per region.
Every method parallelized its permutation walk across all cores of one workstation (Intel i9-14900K, 24 cores / 32 threads) with a single consumer-grade NVIDIA RTX 4060 GPU; the reported wall time is what a user on such a machine waits for.
All methods were tuned, to the author's limited capability, to their best on that machine.
GLOW's wall time was bound by the GPU computing its LLRs: just 10 threads feeding it hierarchical segmentations were enough to saturate its bandwidth.
At the full support GLOW takes about four and a half times as long as the voxel-wise methods.

\begin{figure}[htbp]
	\centering
	\includegraphics[width=.7\textwidth]{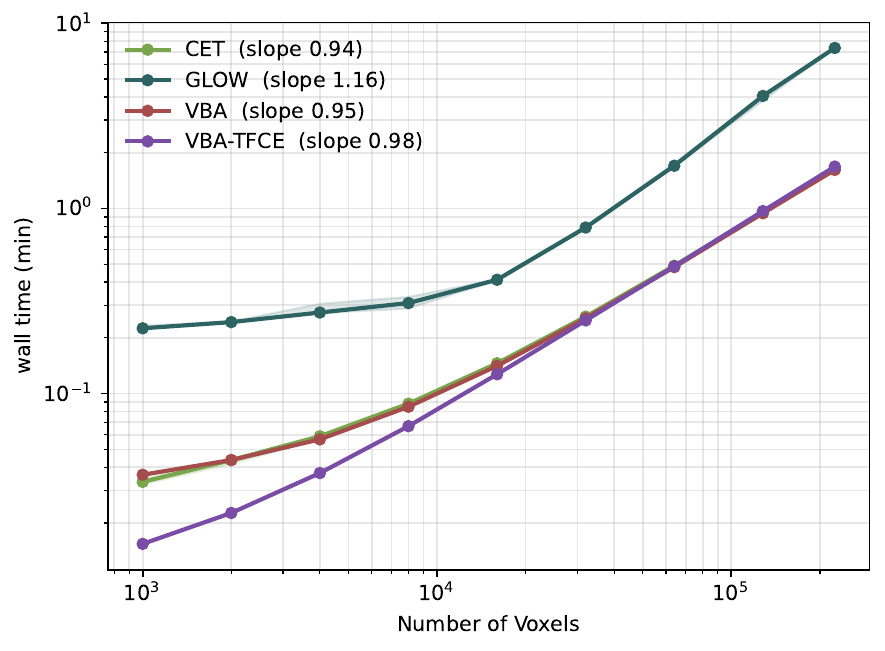}
\caption{Wall-clock fit time versus analyzed voxel count, on log-log axes.
              Each curve is the median over three HCP crops (shaded band: min to max).  Each legend entry gives that method's fitted log-log slope over the largest voxel counts (22{,}462 voxels and up).
			  Note these times use $n_{\mathrm{fwer}} = 500$ outer permutations.}
	\label{fig:runtime}
\end{figure}

Fig~\ref{fig:runtime} shows GLOW scaling roughly linearly with voxel count.
Two things make that possible: the LLR savings above, and constraining segmentation merges to neighboring regions alone (left unconstrained, Ward's method scales as the square of the number of voxels~\parencite{mullner2011modern}).
Extrapolating that linear growth to the full HCP support (225k voxels) at $5{,}000$ permutations, we estimate a GLOW fit at 74 minutes against 16 to 17 minutes for each baseline.

\begin{figure}[htbp]
	\centering
	\includegraphics[width=.8\textwidth]{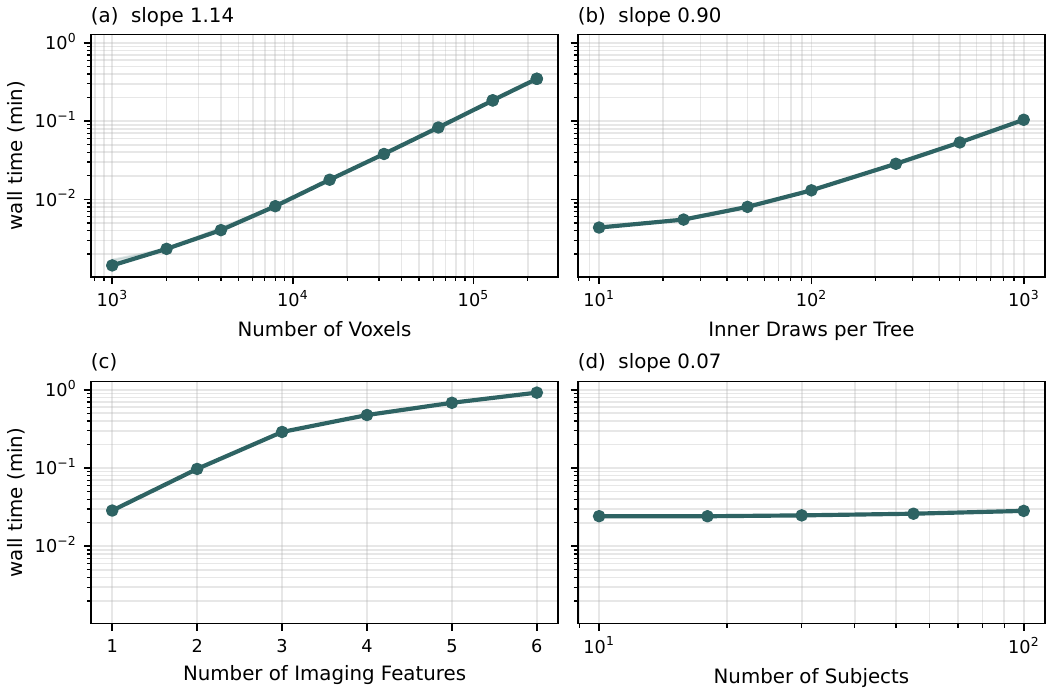}
	\caption{GLOW's per-permutation cost, timed on one core with no GPU and BLAS pinned to one thread.
			  Panels sweep voxel count $|V|$, inner draws per tree $n_{\mathrm{inner}}$, imaging features $b$ and subjects $N$, each holding the rest at the shared center ($25{,}000$ voxels, one outer permutation, $b = 1$, the full cohort).
			  Median over three HCP crops (band: min to max) on a shared wall-time axis; titles give the fitted log-log slope, for $|V|$ over its largest decade as in Fig~\ref{fig:runtime}.}
	\label{fig:runtime_1perm}
\end{figure}

A wall time measured on all cores conflates how the algorithm grows with how well the machine parallelizes it.
To isolate the growth we re-timed GLOW alone on a single pinned core, with no GPU and BLAS held to one thread, at $n_{\mathrm{fwer}} = 1$ (Fig~\ref{fig:runtime_1perm}).

Per permutation, cost grows roughly linearly in voxel count and a little sublinearly in the inner draws, is nearly free in the number of subjects, and rises most steeply in the number of imaging features $b$.

\section{Discussion}\label{sec:discussion}
GLOW's advantage is concentrated where an effect is hard to see, and on real images: on HCP it leads on Dice below $\LLR/|r| = 0.03$ and ties above it, while on white noise it trails from moderate strengths up.
Part of that lead on subtle effects may come from the baselines' kernel width: each runs at the single width that maximizes its mean Dice pooled over effect strengths, while subtle effects favor a wider one (Appendix~\ref{sec:stat_selection}).
A wider kernel would narrow GLOW's lead on subtle effects, but would cost the baselines more Dice on strong effects than it gains on weak effects.
GLOW, by contrast, has no kernel to tune, so it never needs a guess at the effect's strength or geometry in advance.
Compared to voxel-wise baselines, it does a better job handing a spatially consistent effect's whole extent back to the user (the completeness gap of Section~\ref{ssec:result_detection}), at the cost of bundling more tagalong false positive voxels along with it.
GLOW's keystone ingredient is that it reads imaging features across voxels when judging which GLMs are viable.
VBA does great work with per-voxel statistics, but a per-voxel statistic can only characterize so much (Fig~\ref{fig:same_f}).
Where an effect is strong enough to be noticed in each of its voxels alone, VBA does a great job.
Where it is subtle enough to need repeated observation across voxels, GLOW shines.

Detection aside, per-voxel statistics make it hard to construct regions holding all, and only, the voxels of one consistent GLM.
Adjacency among significant voxels may be too permissive a criterion for the job.
Much has been written about a cluster's poor spatial specificity, that its extent licenses no claim about which of its voxels carry the effect \parencite{woo2014cluster,Rosenblatt2018ARI}.
There is a second difficulty: should two distinct effects happen to lie side by side, adjacency alone will not tell them apart.
A sphere planted across a tissue boundary is such a case: a per-voxel test, fitting each voxel its own mean, cannot see the boundary, while GLOW's region LLR can, since pooling the two sides adds their difference in means to $E$ (\ref{eqn:e_partition}).
The Focus objective, which strips each voxel's mean, does not segment along it, however.
A hierarchical segmentation may offer a fruitful vantage point on the problem.

The pruning step presented here takes the common statistical approach of maximizing likelihood, and many of GLOW's troubles trace back to the model generating that likelihood (independent and identically distributed Gaussian residuals):

\begin{equation}
	\LLR = \frac{N|r|}{2} \log \frac{|E + H|}{|E|} \label{eqn:llr_again}
\end{equation}

The $N|r|$ factor encapsulates the repeated measurements of a GLM's deviation from $H_0$; it is the row count of the stacked regression of $Y'$ on $X'$ (\ref{eqn:xy_stacked}).
Given another $N$ images of unrelated subjects showing the same GLM, we would rightly double the LLR.
It is concerning that adding $|r|$ consistent voxels has the same impact.
Correlation across space explains away part of the repeated measure: the extra voxels may only be mimicking their neighbors rather than independently pointing at one underlying GLM.
This raises the essential question: \textbf{how many independent observations does a region of size $|r|$ over $N$ images actually provide?}
TFCE answers implicitly through its $e = 0.5$ and $h = 2$ exponents, trading a signal's spatial extent against its height.
CET does the same with its cluster-forming threshold: higher initial thresholds favor localized effects.
VBA's initial smoothing is clumsy here, requiring one kernel width be chosen a priori even though the optimal width depends on the effect's strength (Appendix~\ref{sec:stat_selection}) and geometry (compact effects prefer wider kernels than branching ones).
Compounding the difficulty, any answer must depend on the image's resolution and spatial autocorrelation, which varies across it, as some anatomy is more consistent than the rest \parencite{hayasaka2003nonstationary}.
Regardless, GLOW is the most naive of all, taking the $N|r|$ count at face value.

At some point a method must decide which is less consistent under $H_0$: an intense GLM observed in a handful of voxels or a subtler GLM observed across many more.
Let us follow this question, and the spatial independence assumption, through GLOW's three steps.

(1) In practice the segmentation caps GLOW's performance on real imaging data: at weak effects on HCP, most of the PPV GLOW leaves on the table lies beyond any pruning rule's reach, because no node of the tree isolates the effect's extent (Fig~\ref{fig:oracle_vs_k}).
Modeling spatial autocorrelation might raise that ceiling.

(2) The studentization of (\ref{eqn:zscore}) allows the hypothesis testing to largely sidestep this question.
Observed regions are compared to permuted versions of themselves which, critically, have the same size and spatial autocorrelation (thanks to block-exchangeability).

(3) Spatial independence has profound consequences for the pruning method.
It makes the likelihood additive across disjoint regions, so no preference is afforded to an intense, localized effect over a diffuse one carrying the same total.
Dynamic Programming feels this as (\ref{eqn:llr_superadd}), where a region's $\LLR$ never exceeds the sum over any partition of it, so the global optimum always fragments; greedy feels it as a raw $\LLR$ which grows with $|r|$, so its argmax overshoots the extent.
GLOW backs away from the clean global optimum in favor of the greedy placeholder only because the second failure is the more forgiving of the two (Section~\ref{ssec:result_pruning}).

A model which captures spatial dependence, such as a conditional autoregressive (CAR) prior \parencite{besag1974spatial}, may improve the segmentation and lift these limits.
An $\LLR$ which priced a region's own spatial dependence might also be comparable across region sizes on its own, retiring the inner permutation loop which dominates GLOW's runtime (Section~\ref{ssec:result_computation}).

\section{Data \& Code Availability}\label{sec:data_code_availability}

GLOW is implemented in Python 3 and is available for use and modification (MIT license) at \weblink{https://github.com/matthigger/glow}{github.com/matthigger/glow}.
The GLOW package would not have been possible without scikit-learn \parencite{scikit-learn}, NumPy \parencite{2020NumPy-Array}, SciPy \parencite{2020SciPy-NMeth}, nibabel \parencite{nibabel}, numba \parencite{lam2015numba}, PyTorch \parencite{ansel2024pytorch} and joblib \parencite{joblib}; numba compiles the segmentation walk and PyTorch runs the GPU $\LLR$ kernels of Section~\ref{ssec:result_computation}.
Figures are drawn with matplotlib \parencite{hunter2007matplotlib} and seaborn \parencite{waskom2021seaborn}.

For validation, we used the 3T diffusion images of the ``100 unrelated subjects'' in the HCP dataset (\weblink{https://balsa.wustl.edu/project?project=HCP_YA}{balsa.wustl.edu}, updated August 2025).
Data were provided by the Human Connectome Project, WU-Minn Consortium (Principal Investigators: David Van Essen and Kamil Ugurbil; 1U54MH091657) funded by the 16 NIH Institutes and Centers that support the NIH Blueprint for Neuroscience Research; and by the McDonnell Center for Systems Neuroscience at Washington University.

Our pre-processing pipeline (\weblink{https://github.com/matthigger/hcp100-dki-noddi-mni/tree/v1.1.0}{github.com/matthigger/hcp100-dki-noddi-mni}, v1.1.0, archived at \weblink{https://doi.org/10.5281/zenodo.20736203}{doi.org/10.5281/zenodo.20736203}) reconstructs six microstructural maps per subject from the HCP diffusion images, orchestrated reproducibly at scale with BABS \parencite{zhao2024babs} over the QSIRecon workflow \parencite{cieslak2021qsiprep}.
DIPY's diffusion-kurtosis fit \parencite{garyfallidis2014dipy,henriques2021dki} (shells with b-value $\le 2000\,\mathrm{s/mm}^2$) yields fractional anisotropy (FA: directional coherence of water diffusion), mean diffusivity (MD: overall diffusion magnitude), and mean kurtosis (MK: non-Gaussianity of diffusion, reflecting microstructural complexity).
AMICO \parencite{daducci2015amico} fits the NODDI model \parencite{zhang2012noddi} to the full multi-shell data, yielding the intra-cellular volume fraction (ICVF: neurite density), the isotropic volume fraction (ISOVF: free-water content), and the orientation dispersion index (OD: angular spread of neurites).
All maps are resampled to a common MNI template, registered with ANTs \parencite{avants2011reproducible,antspyx}.

The resulting image set (\weblink{https://doi.org/10.5281/zenodo.20736221}{doi.org/10.5281/zenodo.20736221}) is available thanks to the \href{https://www.humanconnectome.org/study/hcp-young-adult/document/wu-minn-hcp-consortium-open-access-data-use-terms}{WU-Minn HCP Open Access Data Use Terms}, which permit redistribution of HCP-derived data (provided it is shared under those same terms).
Every experiment given here can be re-generated programmatically; see the instructions at \weblink{https://github.com/matthigger/glow/blob/main/glow/_extra/benchmark/rerun.md}{glow/_extra/benchmark/rerun.md} in the repository.
The provenance records behind every figure and table (one JSON per computed result, holding its inputs, score and timing) are archived at \weblink{https://doi.org/10.5281/zenodo.22664285}{doi.org/10.5281/zenodo.22664285}; the figures and tables redraw from them without the imaging data.
The software also includes a dashboard-style viewer, which shows several regions of the hierarchical segmentation at once alongside their GLM models, their extents overlaid on the original images (\weblink{https://glow-viewer.web.app/}{glow-viewer.web.app}).

\section*{Ethics}
This study is a secondary analysis of de-identified, publicly released data and involved no new data collection.
The Human Connectome Project diffusion data were acquired with Washington University Institutional Review Board approval and participants' written informed consent \parencite{vanessen2013wu}.
Their secondary use here was reviewed by the Northeastern University Institutional Review Board and determined exempt (protocol \#26-05-08, exempt category 4).

\section*{Author Contributions}
All work was carried out by Matt Higger.

\section*{Funding}
This work received no grant funding.

\section*{Declaration of Competing Interests}
The author reports no conflicts of interest.

\section*{Declaration of the use of AI}
Anthropic's Claude \parencite{anthropic2026claude} was used as a coding assistant for portions of the GLOW software implementation and to help draft and revise the language of this manuscript; AI-assisted commits are labeled in the git commit messages.
Claude derived the justification of (\ref{eqn:llr_superadd}) in Appendix~\ref{sec:pool_error_hypothesis} and identified that the candidate MANCOVA statistics are equivalent under VBA and CET (Appendix~\ref{sec:stat_selection}); the author verified both.
No AI system generated the experimental results.
All AI-assisted text and code were reviewed and verified by the author, who takes full responsibility for the content of this work.



\printbibliography

\appendix
\section{Maximized Log-Likelihood of Normal Samples}\label{sec:normal_identity}
Let $Y \in \mathbb{R}^{N \times b}$ represent a set of $N$ vectors drawn from some unknown normal distribution.
The maximum-likelihood estimators $\hat{\mu} = \frac{1}{N} \sum_i y_i$ and $\hat{\Sigma} = \frac{1}{N} \sum_i (y_i - \hat{\mu}) (y_i - \hat{\mu})^T$ provide:
\begin{equation}
	\sum_{i=1}^N (y_i - \hat{\mu})^T \hat{\Sigma}^{-1} (y_i - \hat{\mu})
	= \Tr \sum_{i=1}^N  (y_i - \hat{\mu})(y_i - \hat{\mu})^T \hat{\Sigma}^{-1}
	= N \Tr \hat{\Sigma} \hat{\Sigma}^{-1}
	= Nb
\end{equation}
It follows that the log-likelihood \parencite{bishop2006pattern} of all samples under the fitted distribution is:
\begin{align*}
	\log P(Y|\hat{\mu},  \hat{\Sigma})
	 & =  -\frac{N b}{2} \log(2\pi) - \frac{N}{2} \log|\hat{\Sigma}| - \frac{1}{2} \sum_{i=1}^N (y_i - \hat{\mu})^T \hat{\Sigma}^{-1} (y_i - \hat{\mu})
	\\
	 & =  -\frac{N b}{2} (1 + \log(2\pi)) - \frac{N}{2} \log|\hat{\Sigma}| \numberthis \label{eqn:normal_identity}
\end{align*}

\section{Error and Hypothesis Under Region Pooling}\label{sec:pool_error_hypothesis}
Consider again a region $r$ partitioned into disjoint regions $r_1, \ldots, r_k$, writing $E_r$ and $H_r$ for (\ref{eqn:e_multi_voxel}) and (\ref{eqn:h_multi_vox}) evaluated on $r$.
The mechanism behind (\ref{eqn:llr_superadd}) is visible in the GLM error $E_r$, the total sum of squares, $T_r = \sum_{v \in r} Y_v^T Y_v$, minus the variance explained by $Q$:
\begin{align*}
	E_r & = T_r - |r| \bar{Y}_r^T Q Q^T \bar{Y}_r
	\\
	    & = \sum_i \left( E_{r_i} + |r_i| \bar{Y}_{r_i}^T Q Q^T \bar{Y}_{r_i} \right) - |r| \bar{Y}_r^T Q Q^T \bar{Y}_r
	\\
	    & = \sum_i E_{r_i} + \sum_i |r_i| (\bar{Y}_{r_i} - \bar{Y}_r)^T Q Q^T (\bar{Y}_{r_i} - \bar{Y}_r) \numberthis \label{eqn:e_partition}
\end{align*}
where the second equality uses $T_r = \sum_i T_{r_i}$ and each $T_{r_i}$ is given by the first equality.
The last equality uses $|r| \bar{Y}_r = \sum_i |r_i| \bar{Y}_{r_i}$.
The reduced model's error, $E + H$, also obeys (\ref{eqn:e_partition}) with $Q_0$ in place of $Q$:
\begin{equation}
	E_r + H_r = \sum_i (E_{r_i} + H_{r_i}) + \sum_i |r_i| (\bar{Y}_{r_i} - \bar{Y}_r)^T Q_0 Q_0^T (\bar{Y}_{r_i} - \bar{Y}_r) \label{eqn:e_plus_h_partition}
\end{equation}
Subtracting (\ref{eqn:e_partition}) from (\ref{eqn:e_plus_h_partition}) gives:
\begin{equation}
	H_r = \sum_i H_{r_i} - \sum_i |r_i| (\bar{Y}_{r_i} - \bar{Y}_r)^T Q_1 Q_1^T (\bar{Y}_{r_i} - \bar{Y}_r)
	\label{eqn:h_partition}
\end{equation}
Because $Q Q^T$ and $Q_1 Q_1^T$ are projections, both corrections are positive semi-definite: pooling can only add to the error and only subtract from the hypothesis, with equality when every block's spatial mean agrees with $\bar{Y}_r$ within the column space of $X$.
$\LLR$ (\ref{eqn:llr}) grows with $H$ and shrinks with $E$, so the whole region's ratio $|E_r + H_r| / |E_r|$ cannot exceed the one its parts form together, $|\sum_i (E_{r_i} + H_{r_i})| / |\sum_i E_{r_i}|$.
Reaching (\ref{eqn:llr_superadd}) itself would further require accounting for the weight $N |r_i|$ each part carries in (\ref{eqn:llr}), which we leave aside.

\section{Imposing an Effect: Constant Offset}\label{sec:impose_constant_offset}
For simplicity, we add a constant offset $\Delta \in \mathbb{R}^{N \times b}$ to each $Y_v$ within the region.
Rather than impose an arbitrary effect direction $\beta$,\footnote{Doing so introduces a challenge: $Y$ already carries an incidental effect direction that, without adjustment, would nudge the minimum MSE $\beta$ away from the given $\beta$.  One could correct this by first subtracting the incidental effect, but that only increases $|| \Delta ||_F$ and carries the image set further from its realistic origins.} we scale the existing components of $\bar{Y}_r$ until the region attains the desired effect strength.
Letting $Q_2$ denote any orthonormal basis for the residual space (the orthogonal complement of $X$'s column space, so that $Q_2 Q_2^T = I_N - Q Q^T$), we decompose:
\begin{equation}
	\bar{Y}_r = \sum_{i=0}^2 Q_i Q_i^T \bar{Y}_r \label{eqn:y_decompose_in_q}
\end{equation}
so that our offset is:
\begin{equation}
	\Delta = \sum_{i=0}^2 \alpha_i Q_i Q_i^T \bar{Y}_r
\end{equation}
where $\alpha_i \in \mathbb{R}$.
Orthonormality gives a convenient formulation of our minimization objective:
\begin{equation}
	||\Delta||_F^2 = \sum_{i=0}^2 \alpha_i^2 ||Q_i^T \bar{Y}_r||_F^2  \label{eqn:delta_norm}
\end{equation}

Next, we quantify the constraint that $\Delta$ impose a given $\LLR$.
Substituting $Y_v + \Delta$ for every $Y_v$ modifies the earlier expressions for $E$ (\ref{eqn:e_multi_voxel}) and $H$ (\ref{eqn:h_multi_vox}):
\begin{equation}
	E = S + |r| (1 + \alpha_2)^2 \bar{Y}_r^T Q_2 Q_2^T \bar{Y}_r
	\label{eqn:e_multi_vox_modify}
\end{equation}
\begin{equation}
	H = |r| (1 + \alpha_1)^2 \bar{Y}_r^T Q_1 Q_1^T \bar{Y}_r
	\label{eqn:h_multi_vox_modify}
\end{equation}
from which the $\LLR$ (\ref{eqn:llr}) follows.

Because the offset is constant across space, it has no impact on the spatial covariance $S$ (\ref{eqn:sigma}).
Notice, too, that $\alpha_0$ appears in neither equation above.
The columns of $Q_0$ hold the nuisance variables, whose correlations are projected away before the analysis; offsetting $\Delta$ along those directions only inflates the objective $||\Delta||_F^2$ without moving the $\LLR$, so the optimum sets $\alpha_0 = 0$.

In summary, we choose $\alpha_1$ and $\alpha_2$ to minimize the offset norm (\ref{eqn:delta_norm}) subject to a single constraint: the resulting $\LLR$, computed from (\ref{eqn:e_multi_vox_modify}), (\ref{eqn:h_multi_vox_modify}) and (\ref{eqn:llr}), must match the user-provided target.
We solve this constrained optimization with the Sequential Least Squares Programming (SLSQP) optimizer in SciPy \parencite{2020SciPy-NMeth}.

\section{Tuning the Voxel-wise Baselines}\label{sec:stat_selection}
Each voxel-wise baseline (VBA, VBA-TFCE and CET) carries two choices, its statistic and its smoothing kernel, which we tune jointly: picking one at a time would assume the other's answer.

\paragraph{Statistic.}
Four classical MANCOVA statistics generalize the univariate $F$-statistic: Wilks' Lambda (entered as $1 - \Lambda$, so it grows with evidence against $H_0$), Pillai's Trace, Hotelling--Lawley Trace and Roy's Largest Root \parencite{rencher2012methods}.
The fifth candidate is GLOW's $\LLR$ (\ref{eqn:llr}) on a single-voxel region.
Each also has a z-scored variant, which standardizes every voxel's statistic by its mean and standard deviation across the $n_{\mathrm{fwer}} + 1$ permutation rows, the unpermuted row included \parencite{Phipson2010Permutation}, before the max-statistic null; this puts HCP's spatially heterogeneous voxels on a common scale.

\paragraph{Kernel.}
Before the statistic is computed, every image is smoothed with an isotropic Gaussian of FWHM $0$, $2$, $4$, $8$ or $12$~mm ($0$, $1$, $2$, $4$ or $6$ voxels).
The kernel reads beyond the analyzed crop, and the smoothed images are divided by the identically smoothed brain mask so voxels at the brain's edge are not dimmed.

\paragraph{Selection.}
Each baseline takes the best mean Dice of its 50 variants (5 statistics $\times$ 2 scalings $\times$ 5 kernels) over HCP experiments at $b = 1$: five effect strengths ($0.003$, $0.0075$, $0.03$, $0.12$ and $0.3$) crossed with 10 seeds, other settings at the defaults of Section~\ref{sec:results}.

With one regressor of interest, $H$ (\ref{eqn:h_single_vox}) has rank one, so every candidate is an increasing function of its one non-zero eigenvalue $\theta$: $1 - \Lambda$ and Pillai's Trace equal $\theta / (1 + \theta)$, Hotelling--Lawley and Roy equal $\theta$, and the $\LLR$ is $\frac{N}{2} \log(1 + \theta)$.
An increasing map preserves the ordering of every voxel's statistic, which is all VBA and CET read, so unscaled VBA is exactly invariant to the choice and CET nearly so\footnote{CET's cluster-forming threshold interpolates between two order statistics, and interpolation does not commute with a non-linear map, so a boundary voxel can change sides.  Z-scoring standardizes each voxel after the map and so reorders voxels slightly, which moves VBA's and CET's mean Dice by about $0.001$ (Table~\ref{tab:stat_dice}).}.
VBA-TFCE integrates the statistic's magnitude, and at $2$~mm prefers z-scored Wilks' Lambda (equivalently Pillai's Trace); VBA and CET adopt the same statistic, unscaled.
The preference holds at $b = 2$, where $H$ is still rank one (Table~\ref{tab:stat_dice}).

All three baselines peak at $2$~mm pooled over effect strengths (the mean rows of Table~\ref{tab:fwhm_llr_dice}), but that width wins only two of the fifteen pairings of baseline and strength.
The best width shrinks as effect strength rises, from $4$ to $12$~mm at the weakest effects to no smoothing at the strongest: smoothing buys sensitivity with spatial precision, and a strong effect needs little of the first.
Tuning the width per strength would make each baseline a function of the unknown truth, so each runs at its single best width.

\begin{table}[htbp]
	\centering
\definecolor{shadeVBA}{HTML}{A54C4C}
\definecolor{shadeVBATFCE}{HTML}{794CA5}
\definecolor{shadeCET}{HTML}{79A54C}
\begin{tabular}{llrrrrr}
  \toprule
  \multicolumn{2}{l}{} & \multicolumn{5}{c}{FWHM (mm)} \\
  \cmidrule(lr){3-7}
  Method & LLR$/|r|$ & 0 & 2 & 4 & 8 & 12 \\
  \midrule
  VBA & 0.003 & 0.002 & \cellcolor{shadeVBA!3}0.004 & \cellcolor{shadeVBA!16}0.011 & \cellcolor{shadeVBA!39}0.023 & \cellcolor{shadeVBA!55}\textbf{0.031} \\
   & 0.0075 & 0.031 & \cellcolor{shadeVBA!6}0.041 & \cellcolor{shadeVBA!35}0.087 & \cellcolor{shadeVBA!54}0.116 & \cellcolor{shadeVBA!55}\textbf{0.118} \\
   & 0.03 & 0.242 & \cellcolor{shadeVBA!10}0.277 & \cellcolor{shadeVBA!44}0.393 & \cellcolor{shadeVBA!55}\textbf{0.430} & \cellcolor{shadeVBA!45}0.397 \\
   & 0.12 & \cellcolor{shadeVBA!51}0.848 & \cellcolor{shadeVBA!55}\textbf{0.866} & \cellcolor{shadeVBA!39}0.787 & \cellcolor{shadeVBA!16}0.674 & 0.597 \\
   & 0.3 & \cellcolor{shadeVBA!55}\textbf{0.995} & \cellcolor{shadeVBA!55}0.992 & \cellcolor{shadeVBA!32}0.810 & \cellcolor{shadeVBA!12}0.651 & 0.559 \\
   & mean & 0.424 & \textbf{0.436} & 0.417 & 0.379 & 0.341 \\
  \midrule
  VBA-TFCE & 0.003 & 0.020 & \cellcolor{shadeVBATFCE!26}0.034 & \cellcolor{shadeVBATFCE!55}\textbf{0.051} & \cellcolor{shadeVBATFCE!49}0.047 & \cellcolor{shadeVBATFCE!34}0.039 \\
   & 0.0075 & 0.128 & \cellcolor{shadeVBATFCE!16}0.145 & \cellcolor{shadeVBATFCE!55}\textbf{0.185} & \cellcolor{shadeVBATFCE!39}0.169 & \cellcolor{shadeVBATFCE!23}0.152 \\
   & 0.03 & \cellcolor{shadeVBATFCE!41}0.571 & \cellcolor{shadeVBATFCE!55}\textbf{0.620} & \cellcolor{shadeVBATFCE!54}0.617 & \cellcolor{shadeVBATFCE!29}0.525 & 0.421 \\
   & 0.12 & \cellcolor{shadeVBATFCE!55}\textbf{0.980} & \cellcolor{shadeVBATFCE!54}0.970 & \cellcolor{shadeVBATFCE!30}0.797 & \cellcolor{shadeVBATFCE!11}0.659 & 0.574 \\
   & 0.3 & \cellcolor{shadeVBATFCE!55}\textbf{0.997} & \cellcolor{shadeVBATFCE!51}0.962 & \cellcolor{shadeVBATFCE!25}0.738 & \cellcolor{shadeVBATFCE!9}0.601 & 0.526 \\
   & mean & 0.539 & \textbf{0.546} & 0.478 & 0.400 & 0.342 \\
  \midrule
  CET & 0.003 & 0.027 & \cellcolor{shadeCET!12}0.039 & \cellcolor{shadeCET!42}0.070 & \cellcolor{shadeCET!55}\textbf{0.082} & \cellcolor{shadeCET!43}0.070 \\
   & 0.0075 & 0.120 & \cellcolor{shadeCET!10}0.140 & \cellcolor{shadeCET!51}0.218 & \cellcolor{shadeCET!55}\textbf{0.225} & \cellcolor{shadeCET!33}0.183 \\
   & 0.03 & \cellcolor{shadeCET!17}0.505 & \cellcolor{shadeCET!39}0.567 & \cellcolor{shadeCET!55}\textbf{0.613} & \cellcolor{shadeCET!32}0.546 & 0.454 \\
   & 0.12 & \cellcolor{shadeCET!55}\textbf{0.971} & \cellcolor{shadeCET!54}0.967 & \cellcolor{shadeCET!32}0.806 & \cellcolor{shadeCET!12}0.663 & 0.579 \\
   & 0.3 & \cellcolor{shadeCET!55}\textbf{0.999} & \cellcolor{shadeCET!52}0.978 & \cellcolor{shadeCET!26}0.756 & \cellcolor{shadeCET!9}0.609 & 0.531 \\
   & mean & 0.524 & \textbf{0.538} & 0.492 & 0.425 & 0.363 \\
  \bottomrule
\end{tabular}

	\caption{Mean Dice against kernel width over the 50 tuning experiments (HCP, MinVar extent, $b = 1$), each baseline at its chosen statistic, per effect strength (10 seeds per entry) and pooled over strengths (mean rows).  Bolded entries are each row's maximum.  Each strength row is shaded in its baseline's color, linearly in Dice, from white at the row's minimum to darkest at its maximum.}
	\label{tab:fwhm_llr_dice}
\end{table}

\begin{table}[htbp]
	\centering
\begin{tabular}{llrrrrr}
  \toprule
  Method & Scaling & LLR & Wilks & Pillai & Hotelling & Roy \\
  \midrule
  VBA & raw & 0.411 & 0.411 & 0.411 & \textbf{0.411} & 0.411 \\
   & z-scored & 0.409 & 0.409 & 0.409 & 0.410 & 0.410 \\
  \midrule
  VBA-TFCE & raw & 0.464 & 0.493 & 0.493 & 0.411 & 0.411 \\
   & z-scored & 0.503 & \textbf{0.509} & 0.509 & 0.498 & 0.498 \\
  \midrule
  CET & raw & \textbf{0.492} & \textbf{0.492} & \textbf{0.492} & \textbf{0.492} & \textbf{0.492} \\
   & z-scored & 0.490 & 0.490 & 0.490 & 0.490 & 0.490 \\
  \bottomrule
\end{tabular}

	\caption{Mean Dice over 250 unsmoothed experiments (HCP, MinVar extent, $b = 2$, five effect strengths $\times$ 50 seeds).}
	\label{tab:stat_dice}
\end{table}

\end{document}